\pdfoutput=1

\documentclass[11pt]{article}

\usepackage[final]{acl}

\usepackage{times}
\usepackage{latexsym}
\usepackage{booktabs}
\usepackage{colortbl}
\usepackage{multirow}
\usepackage{subcaption}

\definecolor{highlight}{RGB}{222,238,252} 
\definecolor{lowlight}{RGB}{255,237,222}  
\definecolor{header}{RGB}{240,248,255}
\definecolor{high}{RGB}{0,82,155}  
\definecolor{low}{RGB}{156,51,0}   
\definecolor{neutral}{RGB}{80,80,80}
\definecolor{lightgray}{gray}{0.95}

\usepackage[T1]{fontenc}

\usepackage[utf8]{inputenc}

\usepackage{microtype}

\usepackage{inconsolata}

\usepackage{graphicx}
\usepackage{makecell}
\usepackage{amsmath}

\title{\textsc{Social Caption}:\protect\\Evaluating Social Understanding in Multimodal Models}

\author{Leena Mathur\textsuperscript{*\textdagger}, Bhaavanaa Thumu\textsuperscript{*}, Youssouf Kebe\textsuperscript{*}, Louis-Philippe Morency\\
School of Computer Science, Carnegie Mellon University\\
\texttt{\{lmathur,gyk,morency\}@cs.cmu.edu}}

\begin{document}

\maketitle

\renewcommand{\thefootnote}{\fnsymbol{footnote}}
\footnotetext[1]{Equal contribution \hspace{1em}\textsuperscript{\textdagger}Corresponding author}
\renewcommand{\thefootnote}{\arabic{footnote}}

\begin{abstract}
Social understanding abilities are crucial for multimodal large language models (MLLMs) to interpret human social interactions. We introduce \textsc{Social Caption}, a framework grounded in interaction theory to evaluate social understanding abilities of MLLMs along three dimensions: \textit{Social Inference (SI)}, the ability to make accurate inferences about interactions; \textit{Holistic Social Analysis (HSA)}, the ability to generate comprehensive descriptions of interactions; \textit{Directed Social Analysis (DSA)}, the ability to generate relevant information from interactions. We analyze factors influencing model performance in social understanding, such as scale, architectural design, and spoken context. Experiments with MLLM judges demonstrate a path towards scaling automated evaluation of multimodal social understanding.  

\end{abstract}

\section{Introduction}

Advances in machine learning have enabled multimodal large language models (MLLMs) to perform tasks such as activity recognition and video captioning that require integrating visual, verbal, and vocal information \cite{liang2024hemm, MM-InstructEval}. As MLLMs become deployed in domains that involve social interactions, such as assistive robotics, models must have \textit{social understanding} to operate effectively. This ability is a core component of machine \textit{social intelligence} and involves interpreting multimodal interactions by inferring implicit, often unstated, social information such as the intent, behavior, and interpersonal dynamics of individuals, guided by both group and societal-level  norms \cite{DAUTENHAHN1995333, 5562663, fei2022searching, mathur-etal-2024-advancing}. 

\begin{figure*}[t]
  \centering
  \includegraphics[width=0.93\linewidth]{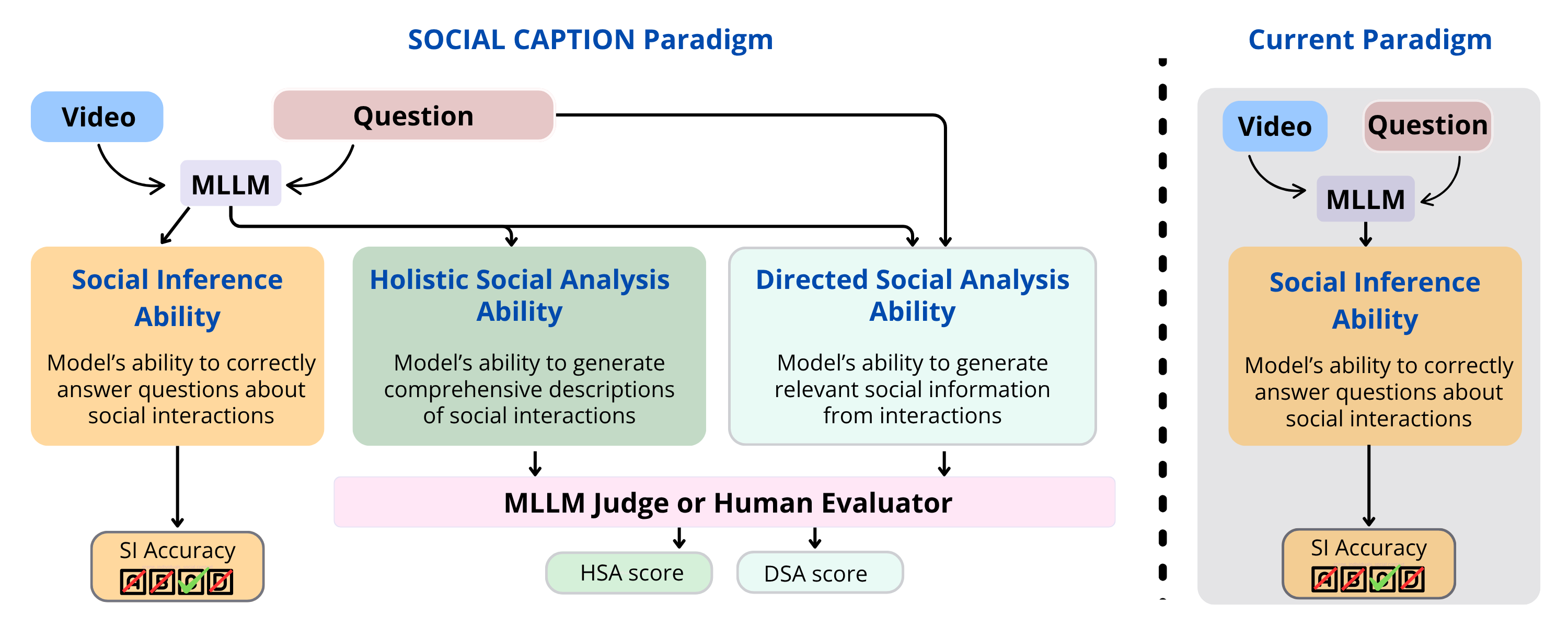}
  \vspace{-0.1cm}
  \caption{\textsc{\textbf{Social Caption}}, a framework for evaluating social understanding of MLLMs, going beyond social inference (\textit{SI}) ability to evaluate holistic social analysis (\textit{HSA}) and directed social analysis (\textit{DSA}) abilities. \textsc{Social Caption} assesses models' understanding of socio-emotional dynamics within multimodal social interactions.}
  \label{fig:evaluation-gap}
  \vspace{-0.3cm}
\end{figure*}

Current approaches to study social understanding of MLLMs use video question answering (QA) to evaluate model  \textit{social inference} abilities  \cite{zadeh2019social, siq2, Grauman_2022_CVPR, kong-etal-2026-siv} 
or measure the quality of model reasoning traces when making QA inferences \cite{mathur-etal-2025-social}. Beyond performing QA (selecting an answer from given options), social understanding involves generating \textit{holistic} interpretations of interactions (e.g., inferring social dynamics from body language) and \textit{directed} interpretations of interactions (e.g., distilling relevant social information towards an inference) \cite{dunn1988beginnings,doi:10.1080/17470218.2014.964738, Azaad_Knoblich_Sebanz_2021}.

We introduce \textbf{\textsc{Social Caption}}, a framework to address the challenge of evaluating social understanding of MLLMs along three dimensions: (1) \textbf{\textit{Social Inference (SI)}}, the ability to make accurate inferences about interactions; (2) \textbf{\textit{Holistic Social Analysis (HSA)}}, the ability to generate comprehensive descriptions of interactions that capture sub-dimensions such as settings, behaviors, and socio-emotional context; (3) \textbf{\textit{Directed Social Analysis (DSA)}}, the ability to generate \textit{relevant} social information from interactions, in response to questions. Figure~\ref{fig:evaluation-gap} shows the difference between current QA-based evaluation for social understanding and the enhanced evaluation enabled by our framework. 

We demonstrate the utility of \textsc{Social Caption} through experiments with state-of-the-art (SOTA) MLLMs and contribute  insights regarding the capabilities and limitations of MLLMs in multimodal social understanding. We contribute novel analysis on factors influencing MLLM social understanding such as scale, architectural design, and presence of spoken context. We experimented with MLLM judges and find that MLLMs demonstrate strong alignment with human scores, introducing a path towards scaling automated evaluation of social understanding in multimodal models. 

\section{Related Work}

\paragraph{Multimodal Social Understanding:} Social understanding refers to the ability to interpret dimensions of social context in interactions. Social context can span the \textit{setting} (e.g., home), \textit{actors} with individual intents, emotions, and goals, the \textit{roles} of these actors (e.g., observer), \textit{social attributes} influencing dynamics (e.g., age), and interaction structures (e.g., dyadic structure) \cite{goffman1959presentation,https://doi.org/10.1111/j.1540-4560.1967.tb00572.x, 5562663, mathur-etal-2024-advancing}. Social understanding involves forming \textbf{holistic} interpretations 
of scenes by synthesizing cues about actors, behaviors, emotions, and intents, as well as making \textbf{directed}, goal-oriented interpretations about interactions \cite{doi:10.1080/17470218.2014.964738,Azaad_Knoblich_Sebanz_2021, 10.3389/fpsyg.2021.743074}.

Prior multimodal social understanding benchmarks such as \textsc{Social-IQ 1.0}, \textsc{Social-IQ 2.0}, \textsc{Ego4D-Social}, and \textsc{SIV-Bench} have measured models' \textit{SI} abilities to answer targeted questions about social interactions, with QA accuracy treated as a proxy for \textit{directed} social reasoning abilities. \textsc{Social Genome} studied the quality of model  reasoning traces during \textit{directed} QA  \cite{zadeh2019social, siq2, Grauman_2022_CVPR, rezaei-etal-2025-egonormia, kong-etal-2026-siv, mathur-etal-2025-social}. \textsc{Social Caption} goes beyond these prior works to introduce a  new evaluation framework for assessing \textit{SI}, \textit{HSA} and \textit{DSA} in multimodal models.

\paragraph{Advancements in MLLMs:} MLLM models typically consist of an LLM backbone, a vision encoder, and a modality alignment module. A common backbone observed in SOTA MLLMs is  Qwen2-7B (including Qwen2-7B-Instruct)~\cite{yang2024qwen2technicalreport}. Some MLLMs utilize SigLIP~\cite{Zhai_2023_ICCV} or CLIP~\cite{pmlr-v139-radford21a} as vision encoders, and others build upon their own trained Vision Transformers (ViTs)~\cite{dosovitskiy2021an} to process image and video. Common alignment strategies include long-context transfer and spatial-temporal modeling~\cite{longva_paper, oryx_paper}, extensive supervised fine-tuning (SFT) on diverse video-text datasets~\cite{chen2024far}, resolution-adaptive techniques~\cite{qwen2vl_paper}, and compact model optimization~\cite{minicpm_paper}. Common SFT strategies include leveraging large-scale video-text datasets to optimize temporal reasoning and multimodal alignment, as well as multi-task learning, instruction tuning, and human-in-the-loop feedback to enhance model performance~\cite{chen2024far, qwen2vl_paper, minicpm_paper}. 
\textsc{Social Caption}  enables researchers to systematically study how MLLM architectural  design decisions translate into multidimensional social understanding abilities.

\paragraph{Automated Evaluation with MLLMs:} The substantial costs and limited scalability of human annotation to evaluate model abilities have led researchers to investigate  approaches for \textit{automated evaluation}. In recent years, researchers have used language models  as proxies for human judgment, with rubrics to evaluate model outputs at scale. This line of research has included text-based frameworks such as \textsc{LLM-Rubric} \cite{hashemi2024llm} and multimodal frameworks such as \textsc{MLLM-as-a-Judge} \cite{3692070.3692324}, \textsc{Judge Anything} \cite{10.1145/3711896.3737409}, and \textsc{MR. Judge} \cite{pi-etal-2025-mr}. MLLM judges can provide reward signals to post-train models  \cite{3692070.3693141, yasunaga2025multimodal, zhang2025videorewardbench}. 

The utility of any MLLM judge relies on the quality of evaluation rubrics \cite{ICLR2026_cfd7bee7} and the validity of the generated scores. \textsc{Social Caption} contributes the first framework to enable  multidimensional evaluation of social understanding, grounded in social interaction theory. We analyze correlations between MLLM judge ratings and human ratings of \textsc{Social Caption} dimensions to contribute insights towards scaling automated MLLM evaluation for social understanding. 

\section{\textsc{Social Caption} Framework}

This section describes the \textsc{Social Caption} evaluation framework (Section \ref{subsec:eval}) and the methodology for human and MLLM judge evaluation (Section \ref{subsec:experimental_setup}). 
A representative example of judge evaluation is in Appendix \ref{sec:appendix_qual}, Figure \ref{fig:qualitative_analysis2}. \textsc{Social Caption} is a dataset-agnostic framework. In this paper, to instantiate this framework, we conduct experiments by augmenting an established testbed, the \textsc{Social-IQ 2.0} dataset~\citep{siq2}, in alignment with prior multimodal social intelligence research \cite{mathur-etal-2025-social, li2025mimeqa}. This testbed  includes real-world interactions paired with a QA task that probes behaviors, intents, emotions, and broader social dynamics of individuals and groups (1 min per video, 145 videos, 943 questions, 4 answer options per question in the validation set).

\subsection{Evaluation Dimensions}
\label{subsec:eval}
\textsc{Social Caption} evaluates 3 distinct, complementary social understanding capabilities in models: Social Inference (\textit{SI}), Holistic Social Analysis (\textit{HSA}), and Directed Social Analysis (\textit{DSA}). Together, these dimensions separate discriminative question answering from open-ended analysis.

\subsubsection{SI Dimension}
The \textit{SI} dimension evaluates a model’s ability to correctly answer multiple-choice questions about social interactions (e.g., \texttt{``Why is the audience laughing?''}). Aligned with prior work \cite{zadeh2019social, siq2, mathur-etal-2025-social, kong-etal-2026-siv}, we use \textit{accuracy} as the evaluation metric. \textit{SI} prompts are in Appendix~\ref{subsec:SI_prompt_template}.
        
\subsubsection{HSA and DSA Dimensions}
 
Our \textit{HSA} and \textit{DSA} evaluation criteria are grounded in the APRACE taxonomy~\cite{10.3389/fpsyg.2021.743074}, a framework derived with grounded theory analysis of 5,000+ open-ended interaction descriptions. We selected APRACE over theoretical frameworks \cite{goffman1959presentation,https://doi.org/10.1111/j.1540-4560.1967.tb00572.x} because it is  derived from observer descriptions of interactions and yields the following components observable from video: \textit{Actors} and \textit{Partners} (attributes and  roles of individuals), \textit{Relations} (structure and dynamics of relationships), \textit{Activities} (behaviors and communicative acts), \textit{Context} (social/temporal setting), and \textit{Evaluation} (scene-level appraisal). We use APRACE to inform the underlying content measured by \textsc{Social Caption} rubrics for \textit{HSA} and \textit{DSA}. \textit{HSA} operationalizes APRACE components in the scene, while \textit{DSA} operationalizes the same components with respect to a question.

\paragraph{Scoring \textit{HSA}}  
Given a video, \textit{HSA} represents a model's ability to generate comprehensive descriptions of social interactions. There are six \textit{HSA} sub-dimensions, bolded below. Four of these sub-dimensions operationalize APRACE components: \textbf{Scene Description} operationalizes the APRACE \textit{Context} component by evaluating the model's description of the setting, time of day, and mood; \textbf{Individuals} operationalizes the APRACE \textit{Actors} and \textit{Partners} components by evaluating appearance, expressions, actions, and body language of people in the scene; \textbf{Topic and Context} operationalizes the APRACE \textit{Activities} and \textit{Relations} components by evaluating the main topics of discussion, background information, and relationship dynamics;  \textbf{Socio-Emotional Analysis} operationalizes the APRACE \textit{Evaluation} component by assessing emotional, power, and conflict/harmony dynamics. The \textit{HSA} rubric additionally includes two sub-dimensions of \textbf{Answer Detail} (amount of detail/nuance in the description) and \textbf{Adherence to Prompt} (extent to which the model followed instructions) to assess answer quality and instruction-following abilities. All six sub-dimensions are rated on a 1-5 Likert scale (maximum \textit{HSA} score of 30). For each video, each MLLM generates a description for \textit{HSA} (details in Appendix \ref{subsec:hsa_rubric}), and this description is scored by human annotators and MLLM judges (details in Appendix \ref{subsec:hsa_mllm_prompt}).  

\paragraph{Scoring \textit{DSA}}
Given a video and a targeted question about interactions in the video, \textit{DSA} represents a model's ability to generate descriptions of relevant social information needed to answer the question. Similar to \textit{HSA}, \textit{DSA} includes six sub-dimensions, bolded below. Four sub-dimensions operationalize the same APRACE components as \textit{HSA}, but evaluate model generations with respect to the question: \textbf{Relevant Scene Details} operationalizes \textit{Context} by evaluating question-relevant setting, time of day, objects, and visual elements; \textbf{Relevant Individuals} operationalizes \textit{Actors} and \textit{Partners} by evaluating question-relevant people, appearances, actions, gestures, and roles; \textbf{Relevant Interactions} operationalizes \textit{Activities} and \textit{Relations} by evaluating interactions between individuals and their significance for the question; \textbf{Relevant Context} operationalizes \textit{Evaluation} by evaluating background information, relationship dynamics, affective cues, and social context needed to answer the question. As done for \textit{HSA}, the \textit{DSA} rubric additionally includes two sub-dimensions of \textbf{Answer Detail} and \textbf{Adherence to Prompt}. All six sub-dimensions are rated on a 1--5 Likert scale (maximum \textit{DSA} score of 30). For each video and corresponding question, each MLLM generates a description for \textit{DSA} (details in Appendix~\ref{subsec:dsa_rubric}), and these descriptions are scored by human annotators and MLLM judges (details in Appendix~\ref{subsec:dsa_mllm_prompt}).

\subsection{Experimental Setup} 
\label{subsec:experimental_setup}
This section describes our  methodology for experiments with \textsc{Social Caption}, covering model selection and human and MLLM judge evaluations.

\subsubsection{Model Selection}
We conducted experiments with a comprehensive set of open-source and closed-source SOTA MLLMs capable of processing videos as input. Models were selected based on performance on video understanding benchmarks such as Video-MME~\citep{11093290} and MMMU~\citep{mmmu}. The open-source models include LongVA~\citep{longva_paper},  Oryx-1.5~\citep{oryx_paper}, LLaVA-NeXT-Video~\citep{llavanextvideo_paper}, MiniCPM-V2.6~\citep{minicpm_paper}, Qwen2-VL~\citep{qwen2vl_paper}, Qwen2.5-VL~\cite{bai2025qwen25vltechnicalreport},  Qwen2.5-Omni~\cite{qwen25omnitechnicalreport}, InternVL2~\citep{Chen_2024_CVPR}, and InternVL3~\cite{zhu2025internvl3}, along with three closed-source models, GPT-4o~\cite{openai2024gpt4ocard}, Gemini-1.5-Pro~\cite{Reid2024Gemini1U} and Gemini-2.5-Pro~\cite{comanici2025gemini}. Appendix \ref{appendix:model_info} and Table \ref{tab:model_info} summarize key modeling design decisions of the open-source MLLMs. We use these models with \textsc{Social Caption} to study how MLLM design decisions influence downstream social understanding abilities. We evaluate \textit{SI} across the broader model set, and conduct the annotation-intensive \textit{HSA} and \textit{DSA} evaluations on the 10 open-source and closed-source models with the highest \textit{SI} performance as a practical screening criterion. 

\subsubsection{Experimental Model Settings}
\label{sec:settings}
Videos were processed at 1 frame per second (FPS) for consistency and sufficient social context. As most open-source models did not support audio input, we omitted the audio modality and  provided WhisperX~\citep{bain23_interspeech} transcriptions with timestamps for temporal dialogue. We used 7B  versions of open-source models when available (three models only had 8B parameters available); we refer to 7-8B parameter models as "standard-scale". 

\subsubsection{SI Evaluations}
To test \textit{SI} for each MLLM, we ran 3 experiments \textit{without} transcriptions   ("w/o trans.") and 3 experiments  \textit{with} transcriptions ("w/ trans."). This setup studies the impact of spoken context on model \textit{SI} abilities. We compute each model's average accuracy for "w/ trans." and "w/o trans." across runs. As all models answer the same questions, \textit{SI} comparisons use paired tests with Holm correction.

\begin{table*}[t]
\centering
\small
\setlength{\tabcolsep}{3.1pt}
\resizebox{\textwidth}{!}{%
\begin{tabular}{@{}l | ccc | cc | cccc | cccc@{}}
\toprule
{\cellcolor{header}\textbf{Model}} &  
{\cellcolor{header}\textbf{Size}} & 
\cellcolor{header}\textbf{CL} &  
{\cellcolor{header}\textbf{Backbone}} & 
\multicolumn{2}{c}{\cellcolor{header}\textbf{Social Inference (SI)}} 
& \multicolumn{4}{c}{\cellcolor{header}\textbf{Holistic Social Analysis (HSA)}}
& \multicolumn{4}{c}{\cellcolor{header}\textbf{Directed Social Analysis (DSA)}}\\
\midrule
\multicolumn{4}{l}{\textbf{Text-Only Base Model}} & 
\textbf{w/o trans. (\%)} & 
\textbf{w/ trans.  (\%)} &
\textbf{H (30)} &
\textbf{G (30)} &
\textbf{I-8B (30)} &
\textbf{I-78B (30)} &
\textbf{H (30)} &
\textbf{G (30)} &
\textbf{I-8B (30)} &
\textbf{I-78B (30)}\\
\midrule
\texttt{Qwen2-7B-Instr.} & 7B & 32K & Qwen2-7B & - & 64.64 & - & - & - & - & - & - & - & -\\
\midrule
\multicolumn{4}{l}{\textbf{Standard-Scale Video-Language Models (7B-8B)}} &
\textbf{w/o trans.  (\%)} & 
\textbf{w/ trans. (\%)} &
\textbf{H (30)} &
\textbf{G (30)} &
\textbf{I-8B (30)} &
\textbf{I-78B (30)} &
\textbf{H (30)} &
\textbf{G (30)} &
\textbf{I-8B (30)} &
\textbf{I-78B (30)}\\
\midrule
\texttt{LongVA} & 7B & 224K & Qwen2-7B-Instr. & \colorbox{lowlight}{52.98}\textcolor{low}{$\downarrow$} & \colorbox{lowlight}{65.43}\textcolor{low}{$\downarrow$} & - & - & - & - & - & - & - & - \\
\texttt{Oryx-1.5} & 7B & 32K & Qwen2.5-7B-Instr. & 56.60 & 67.02 & - & - & - & - & - & - & - & - \\
\texttt{LLaVA-NV} & 7B & 32K & Qwen2-7B & 63.51 & 68.94 & 21.72 & 20.62 & 27.72 & 26.22 & 20.14 & 17.22 & 25.42 & 24.60 \\
\texttt{MiniCPM-V2.6} & 8B & 32K & Qwen2-7B & 63.79 & 69.86 & 22.35 & 22.94 & 29.18 & 27.54 & \textbf{\colorbox{highlight}{23.51}}\textcolor{high}{$\uparrow$} & 22.98 & 27.86 & 26.70\\
\texttt{Qwen2-VL} & 7B & 32K & Qwen2-7B & 65.32 & 72.02 & 21.63 & 20.16 & 26.12 & 24.84 & 18.90 & 16.32 & 23.14 & 22.06\\
\texttt{Qwen2.5-VL} & 7B & 32K & Qwen2.5-7B & 64.26 & 70.00 & \textbf{\colorbox{highlight}{23.24}}\textcolor{high}{$\uparrow$} & \textbf{\colorbox{highlight}{23.62}}\textcolor{high}{$\uparrow$} & 29.26 & 27.48 & 23.49 & \colorbox{highlight}{\textbf{23.16}}\textcolor{high}{$\uparrow$} & 27.68 & 27.16\\
\texttt{Qwen2.5-Omni} & 7B & 32K & Qwen2.5-7B & 63.72 & 68.42 & \colorbox{lowlight}{14.84}\textcolor{low}{$\downarrow$} & \colorbox{lowlight}{13.06}\textcolor{low}{$\downarrow$} & \colorbox{lowlight}{21.74}\textcolor{low}{$\downarrow$} & \colorbox{lowlight}{19.32}\textcolor{low}{$\downarrow$} & \colorbox{lowlight}{8.91}\textcolor{low}{$\downarrow$} & \colorbox{lowlight}{6.88}\textcolor{low}{$\downarrow$} & \colorbox{lowlight}{13.86}\textcolor{low}{$\downarrow$} & \colorbox{lowlight}{11.66}\textcolor{low}{$\downarrow$} \\
\texttt{InternVL2} & 8B & 32K & InternLM2.5-7B & 60.64 & 67.77 & 23.13 & 21.78 & 27.76 & 25.72 & 19.88 & 16.22 & 21.84 & 21.50 \\
\textbf{\texttt{InternVL3}} & \textbf{8B} & \textbf{32K} & \textbf{Qwen2.5-7B} & \colorbox{highlight}{\textbf{67.23}}\textcolor{high}{$\uparrow$} & \colorbox{highlight}{\textbf{73.83}}\textcolor{high}{$\uparrow$} & 21.78 & 23.38 & \colorbox{highlight}{\textbf{29.56}}\textcolor{high}{$\uparrow$} & \colorbox{highlight}{\textbf{28.68}}\textcolor{high}{$\uparrow$} & 23.01 & 22.98 & \colorbox{highlight}{\textbf{28.08}}\textcolor{high}{$\uparrow$} & \colorbox{highlight}{\textbf{27.84}}\textcolor{high}{$\uparrow$}\\
\midrule
\multicolumn{4}{l}{\textbf{Closed-Source Large-Scale Models}} &
\textbf{w/o trans.  (\%)} &
\textbf{w/ trans.  (\%)} &
\textbf{H (30)} &
\textbf{G (30)} &
\textbf{I-8B (30)} &
\textbf{I-78B (30)} &
\textbf{H (30)} &
\textbf{G (30)} &
\textbf{I-8B (30)} &
\textbf{I-78B (30)}\\
\midrule
\texttt{GPT-4o} & - & 128K & - & \colorbox{lowlight}{71.74}\textcolor{low}{$\downarrow$} & 78.58 & \colorbox{lowlight}{22.85}\textcolor{low}{$\downarrow$} & 27.54 & \colorbox{lowlight}{29.16}\textcolor{low}{$\downarrow$} & \colorbox{lowlight}{28.34}\textcolor{low}{$\downarrow$} & 24.42 & 26.70 & 28.28 & 27.50\\
\texttt{Gemini-1.5-Pro} & - & 1M & - & 77.62 & \colorbox{lowlight}{78.40}\textcolor{low}{$\downarrow$} & 24.46 & \colorbox{lowlight}{26.98}\textcolor{low}{$\downarrow$} & 29.78 & 28.38 & \colorbox{lowlight}{23.42}\textcolor{low}{$\downarrow$} & \colorbox{lowlight}{26.42}\textcolor{low}{$\downarrow$} & \colorbox{lowlight}{26.36}\textcolor{low}{$\downarrow$} & \colorbox{lowlight}{24.90}\textcolor{low}{$\downarrow$}\\
\textbf{\texttt{Gemini-2.5-Pro}} & \textbf{-} & \textbf{1M} & \textbf{-} & \textbf{\colorbox{highlight}{79.20}}\textcolor{high}{$\uparrow$} & \textbf{\colorbox{highlight}{82.08}}\textcolor{high}{$\uparrow$} & \colorbox{highlight}{\textbf{26.14}}\textcolor{high}{$\uparrow$} & \colorbox{highlight}{\textbf{29.96}}\textcolor{high}{$\uparrow$} & \colorbox{highlight}{\textbf{30.00}}\textcolor{high}{$\uparrow$} & \colorbox{highlight}{\textbf{29.96}}\textcolor{high}{$\uparrow$} & \colorbox{highlight}{\textbf{25.94}}\textcolor{high}{$\uparrow$} & \colorbox{highlight}{\textbf{29.38}}\textcolor{high}{$\uparrow$} & \colorbox{highlight}{\textbf{29.50}}\textcolor{high}{$\uparrow$} & \colorbox{highlight}{\textbf{28.00}}\textcolor{high}{$\uparrow$} \\
\bottomrule
\end{tabular}
}
\vspace{-0.1cm}
\caption{\textbf{Model Performance in Social Understanding Across \textsc{Social Caption} Dimensions:}
 \textit{Social Inference} (accuracy \%), \textit{Holistic Social Analysis (HSA)} (score out of 30), \textit{Directed Social Analysis (DSA)} (score out of 30). Each model is listed with \textit{Size} (number of parameters), \textit{Context Length (CL)}, and \textit{Backbone} (LLM backbone).  \textit{HSA} and \textit{DSA} are evaluated by humans (H) and by MLLM models as judges: Gemini-2.5-Pro (G), InternVL3-8B (I-8B), and InternVL3-78B (I-78B). Arrows represent \textcolor{high}{$\uparrow$}Highest and \textcolor{low}{$\downarrow$}Lowest performance per category.}
\label{tab:model_performance}
\end{table*}

\subsubsection{HSA and DSA Human Evaluations}
\label{subsec:human_eval_process}
For \textit{HSA} and \textit{DSA} dimensions, we conducted human evaluations of generations from 10 models: the three closed-source models and the top 7 standard-scale models, selected by \textit{SI} performance.

To construct a quality-controlled, representative human evaluation set, we randomly sampled a 50-video subset that closely matched the full validation set across observable factors including SI difficulty, question type, number of people in each social scene, and dialogue density (details in Appendix~\ref{appendix:subset_selection}). This design prioritizes the curation of reliable human judgments while keeping annotation cost feasible. Each annotator watched a video and rated the corresponding \textit{HSA} or \textit{DSA} model generation. Annotators were compensated at  \$15/hr (prorated). The annotation pool was gender-balanced, and each video was evaluated by two English-speaking annotators in the United States with at least a high school education. Evaluations were conducted on the Prolific platform ~\cite{Prolific2024} through an IRB-approved process.

\paragraph{Sample Size and Annotator Agreement} With 10 models generating  \textit{HSA} and \textit{DSA} descriptions for 50 videos, 6 sub-dimensions being rated within each \textit{HSA}
 and \textit{DSA} description, and 2 annotators per video, this study collects 12,000 total human ratings. This sample size balances quality-controlled annotation with statistical power to detect within-video model differences. A paired power analysis indicates that N=50 provides 80\% power to detect effects of approximately $d_{z}$=0.40 (details in Appendix \ref{appendix:sample-comparison}). As each video was annotated by a different pair of annotators, inter-annotator agreement (IAA) was assessed with a binary F1 score, indicating strong reliability in annotation (86.60\% for \textit{HSA} and 84.87\% for \textit{DSA}; details in  Appendix~\ref{appendix:IAA}). To account for repeated evaluation over the same videos, we fit mixed-effects models with model identity as a fixed effect and video as a random intercept, using per-video total scores  across annotators (details in Appendix \ref{appendix:IAA}).  

\subsubsection{MLLM Judges for HSA and DSA}
\label{subsec:mllmeval}
To investigate MLLMs as automated judges for social understanding, we experimented with the highest-performing open-source and closed-source models as evaluators for \textit{HSA} and \textit{DSA}. We provided MLLM judges with the same evaluation instructions as human annotators for consistency. We use Gemini-2.5-Pro, InternVL3-8B, and InternVL3-78B as judges. This set was designed to study evaluator perspectives, as any capable MLLM judge trained on internet-scale data will share architectural or training priors with candidate models. Gemini-2.5-Pro and InternVL3-8B are also candidate models, enabling measurement of self-preference, while InternVL3-78B serves \textit{only} as a judge (excluded from the candidate pool). 

\section{Results and Discussion}
Results from using \textsc{Social Caption} to evaluate multimodal social understanding abilities of models are presented in Table~\ref{tab:model_performance} and discussed below. \textit{SI} results are computed over the full validation set, while \textit{HSA} and \textit{DSA} human evaluation results are computed over the representative N=50 subset. 

\subsection{\textit{SI} Performance Trends}
Table~\ref{tab:model_performance} presents \textit{SI} results. Gemini-2.5-Pro (82.08\% w/ trans.) significantly outperformed all models except for GPT-4o ($p_{\text{adj}}$ = 0.19). Among standard-scale models, InternVL3-8B achieved the highest performance (73.83\% w/ trans.), followed by Qwen2-VL (72.02\% w/ trans.). While no model outperformed the human layperson \textit{SI} accuracy ($\sim$85\%), we find that SOTA MLLMs demonstrate strong zero-shot \textit{SI} abilities. 

\paragraph{Spoken Context and \textit{SI} Ability} Providing transcriptions consistently improved \textit{SI} performance across all models, with accuracy gains up to 13\%. The magnitude of the performance gap between MLLMs with and without spoken context varies by model family. Compared to standard-scale models and GPT-4o, Gemini-1.5-Pro and Gemini-2.5-Pro demonstrate marginal increases in performance with the inclusion of transcripts, indicating that these models have a stronger ability to use non-verbal visual cues to perform \textit{SI} in the no-transcription setting. These findings demonstrate the importance of spoken context in MLLM social reasoning; this modality captures information about intent, emotion, and relationships that informs accurate social inferences \cite{stivers2005introduction}. The text-only baseline model in Table \ref{tab:model_performance} achieves 64.64\%, indicating that spoken context and LLM backbone priors in MLLMs provide strong signals for \textit{SI} performance.

\begin{figure}[t]
  \centering
  \includegraphics[width=1\columnwidth]{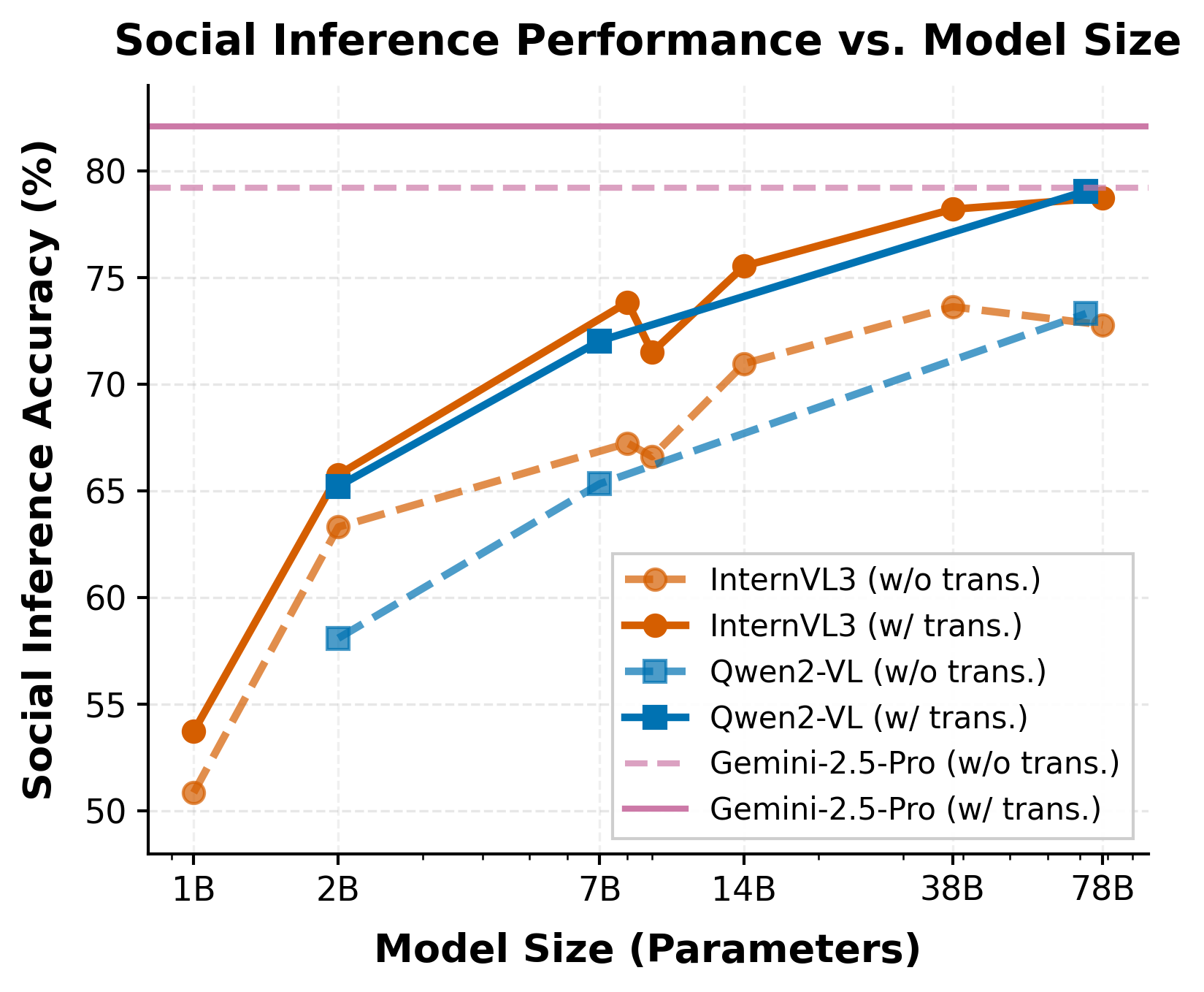}
  \caption{\textbf{Social inference performance across model sizes} with and without spoken context (w/ trans., w/o trans.). Higher scores indicate stronger  capabilities.}
  \label{fig:size_variants}
  \vspace{-0.2cm}
\end{figure}

\paragraph{Model Design and \textit{SI} Ability} As seen in Table \ref{tab:model_performance}, \textit{SI} accuracy varies across MLLM backbones and design decisions. We note that InternVL3 and Qwen2-VL, the highest-performing standard-scale MLLMs, both perform video fine-tuning without any video pre-training. Including video data in pretraining was not found to be a prerequisite for strong downstream \textit{SI} performance. Models vary in the utility of their visual processing for \textit{SI}: LongVA 
does not significantly outperform the text-only Qwen2-7B baseline that uses the same LLM backbone (65.43\% vs 64.64\%, $p_{\text{adj}}$ = 0.25). LongVA treats videos as extended images, with images divided into multiple grids during training \cite{longva_paper}. In contrast, Qwen2-VL uses 3D convolutions and multimodal position embeddings to capture temporal information about inputs \cite{qwen2vl_paper}. Qwen2-VL significantly outperforms both the text-only baseline ($p_{\text{adj}}$ < 0.05) and LongVA ($p_{\text{adj}}$ < 0.01), suggesting that MLLM social reasoning benefits from model architectures that represent videos as continuous streams of temporal input, rather than static sequences of frames. 
   
\paragraph{Model Size and \textit{SI} Ability}  Figure~\ref{fig:size_variants} depicts a controlled analysis to study the impact of model size on \textit{SI} performance for the highest-performing open-source models. Performance results for InternVL3 up to 78B parameters and Qwen2-VL up to 72B parameters are provided in  Appendix \ref{appendix:model_size_variants}, Table \ref{tab:internvl3_qwen2_size_variant}. Larger models typically outperform their smaller counterparts, and models with access to spoken context outperform those without spoken context. The smallest variants InternVL3-1B and Qwen2-VL-2B achieve $\sim$20\% lower accuracy than the largest variants. An exception to this trend was observed for InternVL3-9B, which is larger than InternVL3-8B, but exhibited a lower \textit{SI} performance. We note that all models in the InternVL3 series are built with Qwen2.5 language model backbones \textit{except} for InternVL3-9B, which uses a different backbone (InternLM3). These findings suggest that MLLM architectural decisions and the priors introduced by language model backbones can influence \textit{SI} performance, in addition to scale. 

\textit{The largest variants of InternVL3 and Qwen2-VL achieve }SI\textit{ performance that is competitive with substantially larger closed-source models.} As seen in Table \ref{tab:model_performance} and Figure \ref{fig:size_variants}, there was no significant difference between InternVL3-78B's \textit{SI} performance and that of GPT-4o ($p_{\text{adj}} \geq 1.00$) or Gemini-1.5-Pro ($p_{\text{adj}} \geq 0.56$) and no significant difference between Qwen2-VL-72B's \textit{SI} performance and GPT-4o ($p_{\text{adj}} \geq 1.00$) or Gemini-1.5-Pro ($p_{\text{adj}} \geq 0.25$). 

As seen in Figure~\ref{fig:size_variants}, both InternVL3-78B and Qwen2-VL-72B also approach the \textit{SI} performance of Gemini-2.5-Pro, with a~$\sim$3\% lower accuracy. \textit{Our findings indicate that open-source MLLMs (given careful design decisions, architectures, and post-training strategies) can approach the \textit{SI} capabilities of substantially larger closed-source MLLMs.} These multimodal findings provide empirical evidence that complements prior work in the unimodal domain that demonstrated that scale, alone, is insufficient to elicit social reasoning in models \cite{neural-theory-of-mind}.

\begin{figure}[h]
  \centering
  \includegraphics[width=1\columnwidth]{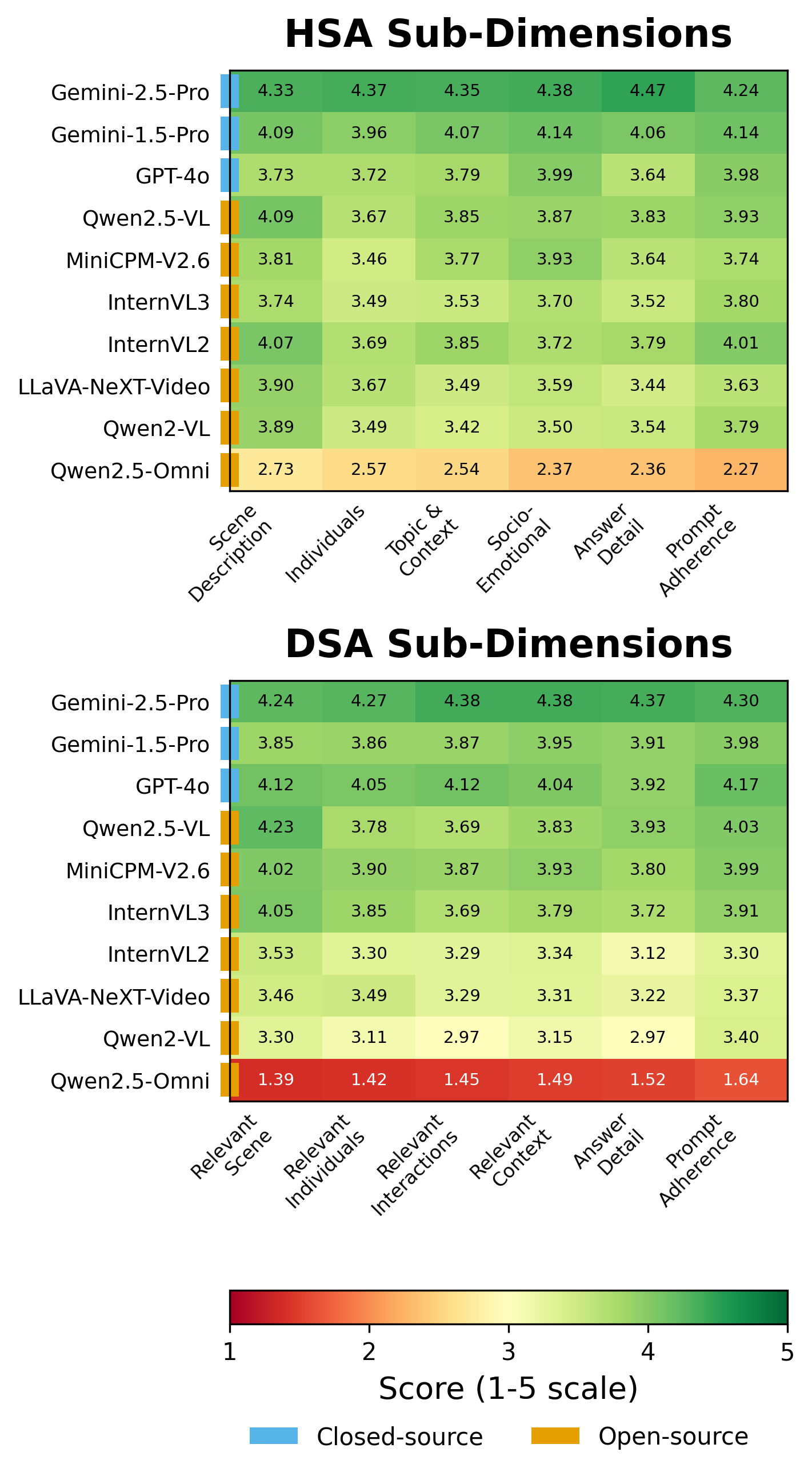}
  \caption{\textbf{\textit{HSA} and \textit{DSA} performance of models across their six evaluation sub-dimensions, visualized as heat maps.} Each cell shows the mean human evaluation score (1–5 scale) for a model–dimension pair, where higher values indicate stronger performance. Color intensity corresponds to score magnitude.}
  \label{fig:heatmap}
  \vspace{-0.3cm}
\end{figure}

\subsection{HSA and DSA Performance}
Table \ref{tab:model_performance} presents \textit{HSA} and \textit{DSA} performance averaged across videos, annotators, and sub-dimensions. HSA and DSA table cells with `-' indicate models that were not included in evaluations (the top 10 MLLMs by \textit{SI} performance were selected for the annotation-intensive \textit{HSA} and \textit{DSA} evaluation). Figure \ref{fig:heatmap} breaks down the human annotator scores across the six \textit{HSA} and \textit{DSA} sub-dimensions.

\paragraph{\textit{HSA} Performance} \textit{HSA} findings from human evaluators revealed that holistic social analysis captures model abilities that are not visible from \textit{SI} accuracy alone. Gemini-2.5-Pro achieved the highest overall \textit{HSA} score (26.14/30). Gemini-2.5-Pro performed significantly better on overall \textit{HSA} than all other evaluated models, including Gemini-1.5-Pro ($p=0.020$) and GPT-4o ($p<0.001$). Among open-source models, Qwen2.5-VL achieved the highest \textit{HSA} score (23.24/30), followed closely by InternVL2 (23.13/30); these two models were statistically indistinguishable. Qwen2.5-VL's performance was also statistically indistinguishable from GPT-4o (22.85/30). These findings suggest that holistic social understanding abilities do not emerge as a result of model size, alone: standard-scale open-source models can approach or match larger closed-source models on \textit{HSA}. 

As seen in Figure \ref{fig:heatmap}, Gemini-2.5-Pro achieved the highest human ratings across all six \textit{HSA} sub-dimensions. Among standard-scale open-source models, Qwen2.5-VL performs highest on \textit{Scene Description}, \textit{Topic and Context}, and \textit{Answer Detail}; InternVL2 performs highest on \textit{Individuals}, \textit{Topic and Context} (tied with Qwen2.5-VL), and \textit{Adherence to Prompt}; MiniCPM-V2.6 performs highest on \textit{Socio-Emotional}. Qwen2.5-Omni struggles across all \textit{HSA} sub-dimensions. These fine-grained findings would be obscured by studying \textit{SI} accuracy alone, but are enabled by  \textsc{Social Caption}.

\paragraph{\textit{DSA} Performance} 
\textit{DSA} findings from human evaluators revealed that Gemini-2.5-Pro achieved the highest  \textit{DSA} score (25.94/30), followed by GPT-4o (24.42/30); this difference was not statistically significant. \textit{DSA} analysis uncovers a tighter comparison between closed-source and open-source models than \textit{SI} accuracy alone suggests. Among open-source models, MiniCPM-V2.6 (23.51/30), Qwen2.5-VL (23.49/30), and InternVL3-8B (23.01/30) achieved the strongest \textit{DSA} scores and were statistically indistinguishable from one another. We did not detect significant differences between MiniCPM-V2.6, Qwen2.5-VL, InternVL3-8B and either GPT-4o or Gemini-1.5-Pro in overall \textit{DSA} performance. These findings suggest that smaller open-source models can generate high-quality descriptions of relevant information in social interactions, on par with generations from larger closed-source models. 

As seen in  Figure~\ref{fig:heatmap}, Gemini-2.5-Pro achieved the highest human ratings across all six \textit{DSA} sub-dimensions. Among open-source models, Qwen2.5-VL performs highest on \textit{Relevant Scene Details}, \textit{Answer Detail}, and \textit{Adherence to Prompt}; MiniCPM-V2.6 performs highest on \textit{Relevant Individuals}, \textit{Relevant Interactions}, and \textit{Relevant Context}. On the sub-dimensions where Qwen2.5-VL and MiniCPM-V2.6 are respectively strongest, their performances are statistically indistinguishable from those of Gemini-2.5-Pro. These findings suggest that standard-scale open-source models can match the performance of larger closed-source models on sub-dimensions of \textit{DSA}.

\paragraph{Model Design and \textit{HSA}/\textit{DSA} Ability} 
We find that stronger performance in \textit{SI} is not a proxy for stronger performance in \textit{HSA} and \textit{DSA}. Qwen2.5-Omni exemplifies this takeaway: this model performs comparably to several standard-scale models in the \textit{SI} task (68.42\%) and has the same Qwen2.5-7B backbone as the Qwen2.5-VL model. However, Qwen2.5-Omni scores substantially lower than Qwen2.5-VL in both HSA (14.84/30 vs 23.24/30) and DSA (8.91/30 vs 23.49/30). This  gap suggests that social analysis abilities in models are shaped by more than the LLM backbone. For example, Qwen2.5-Omni introduces a low-latency streaming architecture for multimodal interaction processing that might have reduced its ability to generate coherent social descriptions of videos. Our findings motivate future MLLM research to study how architecture and streaming algorithms affect the ability of models to perform \textit{HSA} and \textit{DSA}.   

\subsection{MLLM Judges with \textsc{Social Caption}}
\label{subsec:mllmeval_results}

Evaluating candidate model generations of \textit{HSA} and \textit{DSA} with human annotators requires cognitively demanding and time-consuming video annotation \cite{vondrick2013efficiently}. Prior work has demonstrated that MLLM judges can approximate human preferences on open-ended tasks and be used as scalable proxies of human evaluation \cite{3692070.3692324, 10.1145/3711896.3737409}. Whether MLLMs can serve as promising automated judges for multimodal social understanding has not been previously studied. To assess the feasibility of scaling  \textsc{Social Caption} evaluation  with MLLM judges for multimodal social understanding, we studied the extent to which MLLM judges can align with human ratings of \textit{HSA} and \textit{DSA}. We experimented with three judges: Gemini-2.5-Pro, InternVL3-8B, and InternVL3-78B. InternVL3-78B was not among the candidate models and was used only as a judge.
Results are  in Table~\ref{tab:model_performance} and Table ~\ref{tab:binary_agreement}, with details in Appendix \ref{appendix:eval}, Tables~\ref{tab:numbers_hsa} and~\ref{tab:numbers_dsa}.

\paragraph{MLLM Judge Bias} As seen in  Table~\ref{tab:model_performance},   judges ranked Gemini-2.5-Pro the highest in \textit{HSA} and \textit{DSA}. Most judges assigned higher scores than humans did and exhibited mild self-preference effects. These observations of judge behaviors align with LLM judge literature in other domains \cite{chiang2023can, panickssery2024llm}. 
We, therefore, suggest that researchers using \textsc{Social Caption} with MLLM judges primarily consider these scores as relative signals, rather than absolute signals. 

To assess whether MLLM judges are assigning  evaluation scores indiscriminately to model-generated \textit{HSA} and \textit{DSA}  responses, we conduct a \textit{mismatch test}. In this test, each judge evaluates a pair of unrelated video and model response (video-response pair) (details in  Appendix~\ref{sec:baseline_scores}). Across all judges and both \textit{HSA} and \textit{DSA}, mismatch scores remained low, indicating that MLLM judges consistently penalize model generations that do not semantically align with social information in videos.

\paragraph{MLLM Judge Alignment with Human Ratings}
To study the alignment of MLLM judge scores of \textit{HSA} and \textit{DSA} with human scores, we compare human-human (H-H) agreement (across pairs of annotators) to human-judge agreement with the three judge models using a binary F1 score (details in Appendix \ref{appendix:IAA}). Analyses across \textit{HSA} and \textit{DSA} sub-dimensions are in Table \ref{tab:binary_agreement}.

We found strong H-H agreement for both \textit{HSA} and \textit{DSA} 
(average 86.60\% for \textit{HSA} and 84.87\% for \textit{DSA}), indicating that the \textsc{Social Caption} framework can yield reliable annotations from human annotators, despite the inherent subjectivity in assessing social understanding. Human-judge agreement is high across MLLM judges, with F1 scores ranging from 78.81--94.12\% across \textit{HSA} sub-dimensions and 75.24--92.23\% across \textit{DSA} sub-dimensions. 
We note that for each sub-dimension in \textit{HSA} and \textit{DSA}, at least one MLLM judge alignment with human ratings was higher than H-H ratings. These findings indicate that MLLM judges using \textsc{Social Caption} can align with human judgment patterns for \textit{HSA} and \textit{DSA}. 

\begin{table}[t]
\centering
\footnotesize
\renewcommand{\arraystretch}{0.55}
\setlength{\tabcolsep}{2pt}
\begin{tabular}{p{2.25cm}|cccc}
\toprule
\rowcolor{header}\multirow{2}{*}{\cellcolor{header}\textbf{Sub-Dimensions}} & 
    \multicolumn{4}{c}{\cellcolor{header}\textbf{Binary F1 Score (\%)}} \\
\cmidrule(l){2-5}
\rowcolor{header} 
 & \textbf{H-H} & \textbf{H-G} & \textbf{H-I8B} & \textbf{H-I78B} \\
\midrule
\multicolumn{5}{l}{\textbf{HSA}} \\
Scene Desc.      & \colorbox{lowlight}{89.61}\textcolor{low}{$\downarrow$} & 92.80 & 93.93 & \textbf{\colorbox{highlight}{94.12}}\textcolor{high}{$\uparrow$} \\
Individuals        & 83.08 & \colorbox{lowlight}{78.81}\textcolor{low}{$\downarrow$} & 88.84 & \textbf{\colorbox{highlight}{88.91}}\textcolor{high}{$\uparrow$} \\
Topic \& Context       & \colorbox{lowlight}{84.83}\textcolor{low}{$\downarrow$} & 84.95 & 90.97 & \textbf{\colorbox{highlight}{91.42}}\textcolor{high}{$\uparrow$} \\
Socio-Emotional    & 86.89 & \colorbox{lowlight}{85.16}\textcolor{low}{$\downarrow$} & 91.39 & \textbf{\colorbox{highlight}{92.38}}\textcolor{high}{$\uparrow$} \\
Answer Det.        & 85.99 & \colorbox{lowlight}{80.71}\textcolor{low}{$\downarrow$} & 90.42 & \textbf{\colorbox{highlight}{90.46}}\textcolor{high}{$\uparrow$} \\
Prompt Adher.      & \colorbox{lowlight}{88.84}\textcolor{low}{$\downarrow$} & \textbf{\colorbox{highlight}{93.68}}\textcolor{high}{$\uparrow$} & 92.93 & 92.69 \\

\midrule
\multicolumn{5}{l}{\textbf{DSA}} \\
Scene Det. & \colorbox{lowlight}{86.89}\textcolor{low}{$\downarrow$} & \textbf{\colorbox{highlight}{92.13}}\textcolor{high}{$\uparrow$} & 89.81 & 90.91 \\
Individuals  & 84.66 & \colorbox{lowlight}{82.35}\textcolor{low}{$\downarrow$} & 88.86 & \textbf{\colorbox{highlight}{90.15}}\textcolor{high}{$\uparrow$} \\
Rel. Interactions   & 83.91 & \colorbox{lowlight}{78.87}\textcolor{low}{$\downarrow$} & 89.00 & \textbf{\colorbox{highlight}{89.66}}\textcolor{high}{$\uparrow$} \\
Rel. Context     & 85.19 & \colorbox{lowlight}{82.54}\textcolor{low}{$\downarrow$} & 89.31 & \textbf{\colorbox{highlight}{90.50}}\textcolor{high}{$\uparrow$} \\
Answer Det.      & 81.95 & \colorbox{lowlight}{75.24}\textcolor{low}{$\downarrow$} & \textbf{\colorbox{highlight}{89.29}}\textcolor{high}{$\uparrow$} & 89.16 \\
Prompt Adher.    & \colorbox{lowlight}{86.40}\textcolor{low}{$\downarrow$} & \textbf{\colorbox{highlight}{92.23}}\textcolor{high}{$\uparrow$} & 90.72 & 91.12 \\
\bottomrule
\end{tabular}
\caption{\textbf{Binary F1 score (\%) for HSA and DSA Sub-Dimensions.} H-H: Human--Human, H-G: Human--Gemini-2.5-Pro, H-I8B: Human--InternVL3-8B, H-I78B: Human--InternVL3-78B. \textcolor{high}{$\uparrow$} Highest and \textcolor{low}{$\downarrow$} Lowest score per row.}
\label{tab:binary_agreement}
  \vspace{-0.3cm}
\end{table}

We find that human-judge alignment patterns vary across \textit{HSA} and \textit{DSA} sub-dimensions. For \textit{HSA}, InternVL3-78B achieves the highest human-judge alignment on \textit{Scene Description}, \textit{Individuals}, \textit{Topic and Context}, \textit{Socio-Emotional Analysis}, and \textit{Answer Detail}, while Gemini-2.5-Pro achieves the highest human-judge alignment on \textit{Adherence to Prompt}. For \textit{DSA}, InternVL3-78B aligns most strongly with humans on  \textit{Relevant Individuals}, \textit{Relevant Interactions}, and \textit{Relevant Context}; Gemini-2.5-Pro aligns most strongly on \textit{Relevant Scene Details} and \textit{Adherence to Prompt}, and InternVL3-8B aligns most strongly on \textit{Answer Detail}. These findings suggest that the choice of MLLM judge might impact how sub-dimensions of social understanding are evaluated. These experiments demonstrate the potential to use open-source models as evaluators for MLLM social understanding at scale.

\section{Conclusion}
We introduce \textsc{Social Caption}, a new evaluation framework for multimodal social understanding to measure  \textit{Social Inference} (SI), \textit{Holistic Social Analysis} (HSA), and \textit{Directed Social Analysis} (DSA) abilities of MLLMs. We used \textsc{Social Caption} with open-source and closed-source MLLMs to examine their social understanding abilities and factors influencing performance. We find that open-source models can approach \textit{SI} performance of substantially larger closed-source models and can approach or match them on \textit{HSA} and \textit{DSA}. Models with strong \textit{SI} performance do not always have high \textit{HSA} and \textit{DSA} performance. Our experiments with MLLM judges demonstrate the potential of  \textsc{Social Caption} to scale automated evaluation of social understanding. With increasing deployment of MLLMs in AI systems, \textsc{Social Caption} contributes a framework to evaluate and improve social understanding abilities of multimodal models.

\section{Limitations}
\label{sec:limitations}

\paragraph{Length of  Video Interaction Data} All videos used in \textsc{Social Caption} have a length of 1 minute; this experimental setup is consistent with the video lengths used by prior multimodal social intelligence research \cite{lei2018tvqa, zadeh2019social, siq2, mathur-etal-2025-social}. The length of our videos limits the scope of this research and was a design choice for this project; it is not a technical limitation. Our design choice was motivated by prior research establishing that social understanding abilities are important for interpreting short-term, micro-social interactions that occur on the scale of seconds and minutes \cite{beattie1994gestures}.
We note that there remains an open challenge for the research community to study long-context social understanding. Our research motivates future community work to curate longer-form social intelligence video datasets.

\paragraph{Data and Annotation Context}
All videos used by \textsc{Social Caption} have multimodal social interactions in English, and annotators were required to be proficient in English, live in the United States, and have at least a high school education. While this study was not scoped to evaluate social understanding in multilingual or multicultural interactions, 
the \textsc{Social Caption} framework can be used in the future on more diverse video data. Our paper
motivates future community research that builds upon \textsc{Social Caption} to evaluate social understanding abilities in broader contexts. 

\paragraph{Focus and Scope of Paper} \textsc{Social Caption} contributes an evaluation framework and new empirical findings enabled by this framework about MLLM social understanding. Evaluation methodology was our core research focus. Our paper's contributions align with prior machine learning and natural language processing research that treats evaluation as a core research focus.

\section{Ethical Considerations}
\label{sec:ethics}

\paragraph{Ethical Annotation Practices} All video data used by \textsc{Social Caption} is publicly available and released under an academic research license. All annotators were recruited on Prolific and received fair compensation when completing the IRB-approved study (\$15 per hour, pro-rated). The annotation task posed minimal risk to participants, and no personal or identifiable information was collected or stored about annotators, respecting their  confidentiality and privacy. 

\paragraph{Risks for Social Understanding in AI} Our research is intended to contribute towards \textit{societally beneficial} AI systems that can better understand and support humans during interactions. We believe that social understanding abilities in AI systems will be crucial in order for them to effectively work alongside humans to  support human health, well-being, dignity, and autonomy. We note that there are risks to advancing social intelligence in AI, including amplification of social bias \cite{sap2020social}, toxicity \cite{zhou-etal-2023-cobra}, manipulation, and surveillance \cite{10.1145/3544548.3580950, mathur2026should}. We support broader community efforts to mitigate potential
harms of advancing social intelligence in AI systems.

\section*{Acknowledgments}
Leena Mathur acknowledges support by the SoftBank Group-Arm Fellowship and the NSF Graduate Research Fellowship Program under Grant No. DGE2140739. This material is based upon work partially supported by National Institutes of Health awards R1U01MH136535, R01MH125740, R01MH132225, and R21MH130767. Any opinions, findings, conclusions, or recommendations expressed in this material are those of the authors and do not necessarily reflect the views of the sponsors, and no official endorsement should be inferred.

\bibliography{custom}

\clearpage
\appendix

\section*{Appendix}
\section{Qualitative Example}
\label{sec:appendix_qual}

\begin{figure*}[t]
  \centering
  \includegraphics[width=0.97\linewidth]{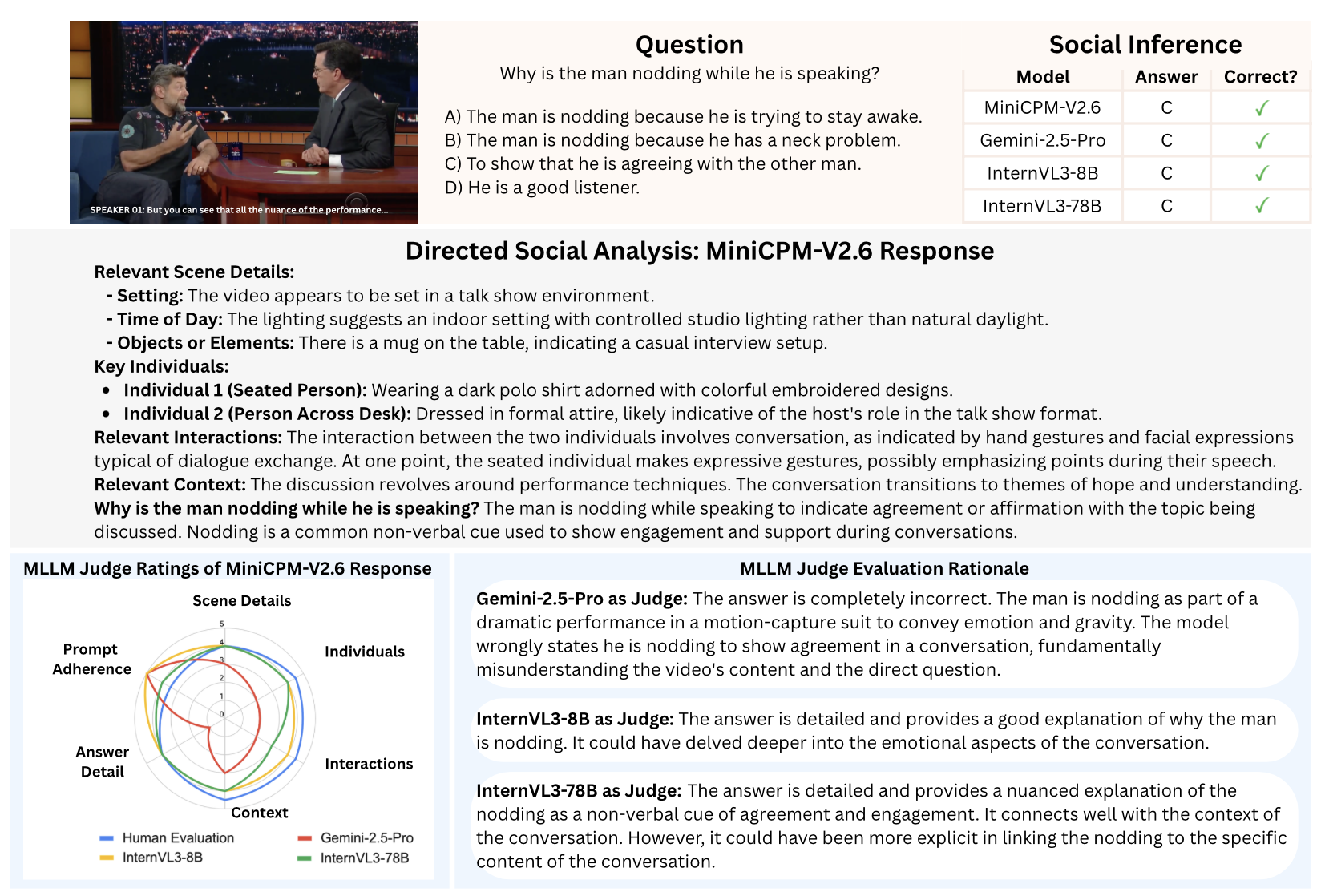}
  \caption{\textbf{Qualitative example of MiniCPM-V2.6 responses evaluated by MLLM judges.} The top section shows a representative video frame, the question, and the SI responses from MiniCPM-V2.6 and MLLM judges. The middle section presents the DSA response of MiniCPM-V2.6, including detailed information on relevant scene details, individuals, interactions and context. The bottom section visualizes human and MLLM judge evaluations of the DSA response in a radar plot, accompanied by judges’ comments explaining reasoning.}
    \label{fig:qualitative_analysis2}
      \vspace{-0.1cm}
\end{figure*}

Figure~\ref{fig:qualitative_analysis2} illustrates a representative example of MLLM judge behaviors. We use an example with  MiniCPM-V2.6, a  standard-scale model that was not used as a judge model. In this example, MiniCPM-V2.6 and the MLLM judges (Gemini-2.5-Pro, InternVL3-8B and InternVL3-78B) have equal performance in \textit{SI} by answering the given question correctly. However, they differ substantially in  evaluation of the corresponding \textit{DSA} generation. Gemini-2.5-Pro assigns a low score to the \textit{DSA} generation across sub-dimensions and deems the response "completely incorrect", while both InternVL judges score it favorably with critiques that are broadly consistent with the human assessment. This discrepancy mirrors the quantitative findings on human-judge alignment in Table \ref{tab:binary_agreement}. Compared to InternVL models, Gemini-2.5-Pro has weaker alignment with humans when assessing sub-dimensions such as information about individuals, interactions, and social context. This qualitative case study illustrates how \textsc{Social Caption} enables fine-grained analysis of models' social understanding abilities.

\section{\textit{SI} Prompts}
\label{subsec:SI_prompt_template}
We provide  prompt templates used to evaluate the \textit{SI} dimension by asking the model to answer multiple-choice questions. Two variants of each \textit{SI} prompt were used: without video transcriptions ("w/o trans.") and  with video transcriptions ("w/ trans.").

\paragraph{\textit{SI}  Prompts Without Transcriptions}
For this variant, the model is required to select the correct answer based \textit{solely} on the video content. The prompt includes an in-context example to illustrate the expected format for the answer. Figure~\ref{fig:social_inference_prompt_without_transcription} presents an example of the prompt.

\begin{figure}[h]
\centering
\definecolor{promptbg}{RGB}{245,245,245}
\setlength{\fboxsep}{6pt}
\setlength{\parskip}{1pt}
\fcolorbox{black}{promptbg}{
\parbox{0.95\columnwidth}{
\scriptsize
\ttfamily
Given the provided video clip of a social interaction, answer the multiple-choice question.\\

\textit{Example from another video:}

\textbf{Question:} How does the woman in black feel about the woman in the dress?

\textbf{Options:}  \\
A. She is interested as she looks her way.  \\
B. The woman in black is a fan of the Crazy Trilogy movies.  \\
C. She acts indifferent, but is actually attracted to her.  \\
D. The woman in black is annoyed by the woman in the dress.  

\textbf{Correct Answer:} D \\

Now, please answer this question about the given video. Please provide only the letter of the answer (e.g. A, B, C, D) and nothing else: \\

\textbf{Question:} \{Question\}

\textbf{Options:}  \\
A. \{Choice A\}  \\
B. \{Choice B\}  \\
C. \{Choice C\}  \\
D. \{Choice D\}  

\textbf{Correct Answer:}
}}
\caption{Prompt template to evaluate the \textit{SI} dimension without providing transcriptions.}
\label{fig:social_inference_prompt_without_transcription}
\end{figure}

\paragraph{\textit{SI} Prompts With Transcriptions}
For this variant, the model is provided with the video and spoken content in the form of transcriptions with timestamps and speaker turns. This additional context allows the model to use dialogue cues, in addition to visual information, to answer a given question. The prompt includes an in-context example to illustrate the expected format for the answer. Figure~\ref{fig:social_inference_prompt_with_transcription} presents an example of the prompt. 

\begin{figure}[b!]
\centering
\definecolor{promptbg}{RGB}{245,245,245} 
\setlength{\fboxsep}{6pt} 
\setlength{\parskip}{1pt} 
\fcolorbox{black}{promptbg}{
\parbox{0.95\columnwidth}{
\scriptsize
\ttfamily
Given the provided video clip of a social interaction and its transcription, answer the multiple-choice question. \\

Example from another video:\\
\textbf{Question:} How does the woman in black feel about the woman in the dress? \\
\textbf{Options:}  \\
A. She is interested as she looks her way.  \\
B. The woman in black is a fan of the Crazy Trilogy movies.  \\
C. She acts indifferent, but is actually attracted to her.  \\
D. The woman in black is annoyed by the woman in the dress.  \\
\textbf{Correct Answer:} D \\

Now, please answer this question about the given video and the transcription. Please provide only the letter of the answer (e.g., A, B, C, D) and nothing else. \\

\textbf{Transcription:} \{Transcription\}  \\
\textbf{Question:} \{Question\}  \\
\textbf{Options: } \\
A. \{Choice A\}  \\
B. \{Choice B\}  \\
C. \{Choice C\}  \\
D. \{Choice D\} \\
\textbf{Correct Answer:}
}}
\caption{Prompt template to evaluate  the \textit{SI} dimension with  transcriptions.}
\label{fig:social_inference_prompt_with_transcription}
\end{figure}

\section{\textit{HSA} and \textit{DSA} Prompts}

\subsection{\textit{HSA} Generations}
\label{subsec:hsa_rubric}
To obtain \textit{HSA} generations, models were provided with each video and corresponding transcription. Models were instructed to generate a description of the scene, individuals, topic \& context, and socio-emotional information. The prompt to obtain \textit{HSA} generations for each video is in Figure~\ref{fig:hsa_prompt}.

\begin{figure}[t]
\centering
\definecolor{promptbg}{RGB}{245,245,245} 
\setlength{\fboxsep}{6pt} 
\setlength{\parskip}{1pt} 
\fcolorbox{black}{promptbg}{
\parbox{0.95\columnwidth}{
\scriptsize
\ttfamily
Please watch the provided video and provide a detailed description of the scene, focusing on the socio-emotional context, dialogue, and topic. Include the following information in your description: \\

\textbf{Transcription:} \{Transcription\} \\

\textbf{1. Scene Description: } \\
- \textbf{Setting:} Describe the location and environment where the video takes place.  \\
- \textbf{Time of Day:} Mention the time of day if it is evident from the video (e.g., morning, afternoon, evening).  \\
- \textbf{Mood:} Describe the overall mood or atmosphere of the scene (e.g. cheerful, tense, relaxed). \\ 

\textbf{2. Individuals:}  \\
For each person in the video, provide the following details:  \\
- \textbf{Appearance:} Describe their physical appearance, including age, gender, ethnicity, clothing, and distinctive features. \\ 
- \textbf{Facial Expressions:} Describe any notable facial expressions or emotions displayed by the individual throughout the video.  \\
- \textbf{Actions:} Describe any significant actions, gestures, or body language exhibited by the individual.  \\

3. \textbf{Topic and Context:}  \\
- \textbf{Main Topic:} Identify the main topic or subject of discussion in the video.  \\
- \textbf{Context:} Provide any relevant context or background information necessary to understand the conversation.  \\
- \textbf{Relationship Dynamics:} Describe the apparent relationship between the individuals and how it influences their interaction.  \\

4. \textbf{Socio-Emotional Analysis:  } \\
- \textbf{Emotional Dynamics:} Analyze the emotional dynamics between the individuals throughout the video. \\
- \textbf{Power Dynamics:} Describe any evident power dynamics or social hierarchy among the individuals.  \\
- \textbf{Conflict or Harmony:} Mention if there are any instances of conflict, disagreement, or harmony in the interaction.  \\

Please provide the description in a structured format, using headings and subheadings for clarity. The description should be comprehensive and capture the nuances of the social interaction and emotional context portrayed in the video.
}}
\caption{\textbf{Prompt template for  \textit{HSA} generations.} The model generates a detailed description of the video using both visual content and spoken dialogue, guided by the structured elements specified in the prompt.}
\label{fig:hsa_prompt}
\end{figure}

\subsection{Human and MLLM Judge \textit{HSA} Prompt}
\label{subsec:hsa_mllm_prompt}
The prompt template used for the human evaluator and MLLM judges to evaluate model generations along the \textit{HSA} dimension is in Figure ~\ref{fig:hsa_mllm_prompt}. The judge receives the model-generated description and the video transcription and is asked to score the response according to a structured rubric (6 items, each rated on a Likert scale 1-5, maximum total score of 30). 

\begin{figure*}[t]
\centering
\definecolor{promptbg}{RGB}{245,245,245} 
\setlength{\fboxsep}{6pt} 
\setlength{\parskip}{1pt} 
\fcolorbox{black}{promptbg}{
\parbox{0.9\textwidth}{
\scriptsize
\ttfamily
\tiny
\textbf{Task:} You are an evaluator evaluating a vision-language model's response to a video analysis prompt. \\

\textbf{Instructions on the evaluation:} \\
1. First, watch the provided video carefully to understand the actual content, context, and details  \\
2. Read the model's response below \\
3. Compare the model's description against what you actually observe in the video \\
4. Rate the model's response using the 6 criteria and 5-point scale provided \\

\textbf{\# Original prompt to the vision-language model STARTS here \# }

Please watch the provided video and provide a detailed description of the scene, focusing on the socio-emotional context, dialogue, and topic. Include the following information in your description: \\

\textbf{Transcription:} \{Transcription\} \\

\textbf{1. Scene Description: } \\
- \textbf{Setting:} Describe the location and environment where the video takes place.  \\
- \textbf{Time of Day:} Mention the time of day if it is evident from the video (e.g., morning, afternoon, evening).  \\
- \textbf{Mood:} Describe the overall mood or atmosphere of the scene (e.g. cheerful, tense, relaxed). \\ 

\textbf{2. Individuals:}  \\
For each person in the video, provide the following details:  \\
- \textbf{Appearance:} Describe their physical appearance, including age, gender, ethnicity, clothing, and distinctive features. \\ 
- \textbf{Facial Expressions:} Describe any notable facial expressions or emotions displayed by the individual throughout the video.  \\
- \textbf{Actions:} Describe any significant actions, gestures, or body language exhibited by the individual.  \\

3. \textbf{Topic and Context:}  \\
- \textbf{Main Topic:} Identify the main topic or subject of discussion in the video.  \\
- \textbf{Context:} Provide any relevant context or background information necessary to understand the conversation.  \\
- \textbf{Relationship Dynamics:} Describe the apparent relationship between the individuals and how it influences their interaction.  \\

4. \textbf{Socio-Emotional Analysis:  } \\
- \textbf{Emotional Dynamics:} Analyze the emotional dynamics between the individuals throughout the video. \\
- \textbf{Power Dynamics:} Describe any evident power dynamics or social hierarchy among the individuals.  \\
- \textbf{Conflict or Harmony:} Mention if there are any instances of conflict, disagreement, or harmony in the interaction.  \\

Please provide the description in a structured format, using headings and subheadings for clarity. The description should be comprehensive and capture the nuances of the social interaction and emotional context portrayed in the video. \\

\textbf{\# Original prompt to the vision-language model ENDS here \#} \\

\textbf{Original model's response to evaluate:} \{Model\_response\} \\

\textbf{Evaluation Rubric} \\
Each response will be rated on a five-point Likert scale across several criteria. Please follow these instructions for each video. For each response, rate it based on the following criteria:\\
\textbf{1. Scene Description (1-5):} Does the response provide a good description of the setting, time of day, and mood compared to what you see in the video?\\
\textbf{2. Individuals (1-5):} Does the response describe the appearance, facial expressions, and actions of the individuals accurately compared to what you see in the video?\\
\textbf{3. Topic and Context (1-5):} Does the response capture the main topic and provide relevant context, including relationship dynamics compared to what you see in the video?\\
\textbf{4. Socio-Emotional Analysis (1-5):} How well does the response analyze emotions, power dynamics, and conflicts or harmony in the scene compared to what you see in the video?\\
\textbf{5. Answer Detail (1-5):} Is the description detailed and nuanced, offering in-depth social and emotional insights compared to what you see in the video?\\
\textbf{6. Adherence to the Prompt (1-5):} Does the description follow the instructions and is the description following the answer structure provided in the prompt compared to what you see in the video?\\

For each of these categories, rate from 1 (Poor) to 5 (Excellent), with a total possible score of 30 points for each description, based on the instructions in the prompt. \\

\textbf{Rating Scale:} \\
\textbf{1 (Poor):} Lacks detail or is incorrect.  \\
\textbf{2 (Fair):} Partially accurate but lacks depth. \\ 
\textbf{3 (Satisfactory):} Provides adequate information, but could be improved.  \\
\textbf{4 (Good):} Largely accurate, with minor omissions.  \\
\textbf{5 (Excellent):} Comprehensive, detailed, and captures nuances perfectly. \\

\textbf{Required Output Format:} \\
\{\{"scene\_description": [score],\\
"scene\_description\_reason": "[brief explanation comparing model response to actual video]",\\
"individuals": [score], \\
"individuals\_reason": "[brief explanation comparing model response to actual video]",\\
"topic\_and\_context": [score],\\
"topic\_and\_context\_reason": "[brief explanation comparing model response to actual video]",\\
"socio\_emotional\_analysis": [score],\\
"socio\_emotional\_analysis\_reason": "[brief explanation comparing model response to actual video]",\\
"answer\_detail": [score],\\
"answer\_detail\_reason": "[brief explanation of response depth and insight quality]",\\
"adherence\_to\_prompt": [score],\\
"adherence\_to\_prompt\_reason": "[brief explanation of how well response follows requested structure]"\}\}\\

\textbf{IMPORTANT Remember:} Base your ratings on how well the model's response matches what you actually see and hear in the video, not on how well-written the response sounds.\\

\textbf{Your Output:} 
}}
\caption{Instructions for Humans and MLLM Judges to evaluate model generations on \textit{HSA} sub-dimensions.}
\label{fig:hsa_mllm_prompt}
\end{figure*}

\subsection{\textit{DSA}  Generations}
 \label{subsec:dsa_rubric}
To obtain \textit{DSA} generations,  models were provided with each video and corresponding transcription. Given a question about the social interaction in the video, models were instructed to generate a description of the scene with relevant details necessary to answer the given question. The prompt to obtain \textit{DSA} generations for each video and corresponding question is in Figure ~\ref{fig:dsa_prompt}.

\begin{figure}[h]
\centering
\definecolor{promptbg}{RGB}{245,245,245} 
\setlength{\fboxsep}{6pt} 
\setlength{\parskip}{1pt} 
\fcolorbox{black}{promptbg}{
\parbox{0.95\columnwidth}{
\scriptsize
\ttfamily
Please watch the provided video and generate a description that helps answer the given question. In your description, focus on the elements of the video that are most relevant to answering the question. Include the following information: \\

\textbf{Transcription:} \{Transcription\} \\

1. \textbf{Relevant Scene Details:}  \\
- \textbf{Setting:} Describe the location and environment relevant to the question.  \\
- \textbf{Time of Day:} Mention the time of day if it's pertinent to the question.  \\
- \textbf{Objects or Elements:} Describe any objects, props, or visual elements that are crucial to answering the question.  \\

2. \textbf{Key Individuals:}  \\
For each person in the video who is relevant to the question, provide the following details:  \\
- \textbf{Appearance:} Describe their physical appearance, focusing on aspects that relate to the question.  \\
- \textbf{Actions:} Describe any actions, gestures, or body language that are significant in the context of the question.  \\

3. \textbf{Relevant Interactions:  } \\
- Describe any interactions between individuals that are pertinent to answering the question.  \\
- Analyze the nature and significance of these interactions in relation to the question.  \\

4. \textbf{Relevant Context:  } \\
- Provide any background information or context from the video that is necessary to understand and answer the question.  \\
- Explain how this context relates to the question and contributes to the answer.  \\

Please structure your description using headings and subheadings to clearly organize the information. Ensure that the description focuses on the aspects of the video that are most relevant to answering the following question. \\

\textbf{Question:} \{Question\}
}}
\caption{\textbf{Prompt template for \textit{DSA} generations.} The model generates a detailed description of relevant information in the video, guided by the structured elements specified in the prompt.}
\label{fig:dsa_prompt}
\end{figure}

\subsection{Human and MLLM Judge \textit{DSA} Prompt}
\label{subsec:dsa_mllm_prompt}

The prompt template used for the human evaluator and MLLM judges to evaluate model generations along the \textit{DSA} dimension is in Figure ~\ref{fig:mllm_dsa_prompt}. The judge receives the model-generated response, the video transcription, and the question about the video and is asked to score the response according to a structured rubric (6 items, each rated on a Likert scale 1-5, maximum total score of 30).

\begin{figure*}[t]
\centering
\definecolor{promptbg}{RGB}{245,245,245} 
\setlength{\fboxsep}{6pt} 
\setlength{\parskip}{1pt} 
\fcolorbox{black}{promptbg}{
\parbox{0.90\textwidth}{
\scriptsize
\ttfamily
\tiny
\textbf{Task:} You are an evaluator evaluating a vision-language model's response to a video analysis prompt. \\

\textbf{Instructions on the evaluation:} \\
1. First, watch the provided video carefully to understand the actual content, context, and details  \\
2. Read the model's response below \\
3. Compare the model's description against what you actually observe in the video \\
4. Rate the model's response using the 6 criteria and 5-point scale provided \\

\textbf{\# Original prompt to the vision-language model STARTS here \# }

Please watch the provided video and generate a description that helps answer the given question. In your description, focus on the elements of the video that are most relevant to answering the question. Include the following information: \\

\textbf{Transcription:} \{Transcription\} \\

1. \textbf{Relevant Scene Details:}  \\
- \textbf{Setting:} Describe the location and environment relevant to the question.  \\
- \textbf{Time of Day:} Mention the time of day if it's pertinent to the question.  \\
- \textbf{Objects or Elements:} Describe any objects, props, or visual elements that are crucial to answering the question.  \\

2. \textbf{Key Individuals:}  \\
For each person in the video who is relevant to the question, provide the following details:  \\
- \textbf{Appearance:} Describe their physical appearance, focusing on aspects that relate to the question.  \\
- \textbf{Actions:} Describe any actions, gestures, or body language that are significant in the context of the question.  \\

3. \textbf{Relevant Interactions:  } \\
- Describe any interactions between individuals that are pertinent to answering the question.  \\
- Analyze the nature and significance of these interactions in relation to the question.  \\

4. \textbf{Relevant Context:  } \\
- Provide any background information or context from the video that is necessary to understand and answer the question.  \\
- Explain how this context relates to the question and contributes to the answer.  \\

Please structure your description using headings and subheadings to clearly organize the information. Ensure that the description focuses on the aspects of the video that are most relevant to answering the following question. \\

\textbf{Question:} \{Question\} \\

\textbf{\# Original prompt to the vision-language model ENDS here \#} \\

\textbf{Original model's response to evaluate:}
\{Model\_response\}  \\

\textbf{Evaluation Rubric}\\
Each response will be rated on a five-point Likert scale across several criteria. Please follow these instructions for each video. For each answer, rate it based on the following criteria:\\
\textbf{1. Relevant Scene Details (1-5):} Does the answer accurately describe the setting, time of day, and any objects that are crucial for answering the question compared to what you see in the video?\\
\textbf{2. Key Individuals (1-5):} Does the answer provide detailed and relevant descriptions of the individuals’ appearance, actions, and gestures that are significant to the question compared to what you see in the video?\\
\textbf{3. Relevant Interactions (1-5):} Does the answer capture and analyze the interactions between individuals that are most pertinent to answering the question compared to what you see in the video?\\
\textbf{4. Relevant Context (1-5):} Does the answer include necessary background or contextual information from the video that contributes to answering the question compared to what you see in the video?\\
\textbf{5. Answer Detail (1-5):} Is the answer detailed and nuanced, offering in-depth social and emotional insights compared to what you see in the video?\\
\textbf{6. Adherence to the Prompt (1-5):} Does the answer follow the instructions and is the answer following the structure provided in the prompt compared to what you see in the video?\\

For each of these categories, rate from 1 (Poor) to 5 (Excellent), with a total possible score of 30 points for each description, based on the instructions in the prompt. \\

\textbf{Rating Scale:} \\
\textbf{1 (Poor):} Lacks detail or is incorrect.  \\
\textbf{2 (Fair):} Partially accurate but lacks depth.  \\
\textbf{3 (Satisfactory):} Provides adequate information, but could be improved.  \\
\textbf{4 (Good):} Largely accurate, with minor omissions.  \\
\textbf{5 (Excellent):} Comprehensive, detailed, and captures nuances perfectly. \\

\textbf{Required Output Format:} \\
\{\{"relevant\_scene\_details": [score], \\
"relevant\_scene\_details\_reason": "[brief explanation comparing model's scene description to actual video setting, time, and objects]", \\
"key\_individuals": [score],\\
"key\_individuals\_reason": "[brief explanation comparing model's individual descriptions to actual people in video]",\\
"relevant\_interactions": [score],\\
"relevant\_interactions\_reason": "[brief explanation comparing model's interaction analysis to actual interactions in video]",\\
"relevant\_context": [score],
"relevant\_context\_reason": "[brief explanation comparing model's contextual information to actual video context]",\\
"answer\_detail": [score],\\
"answer\_detail\_reason": "[brief explanation of response depth and insight quality compared to video content]",\\
"adherence\_to\_prompt": [score],\\
"adherence\_to\_prompt\_reason": "[brief explanation of how well response follows requested structure and instructions]"\}\}\\

\textbf{IMPORTANT Remember:} Base your ratings on how well the model's response matches what you actually see and hear in the video, not on how well-written the response sounds.\\

\textbf{Your Output:}
}}
\caption{Instructions for Humans and MLLM Judges to evaluate model generations on \textit{DSA} sub-dimensions.}
\label{fig:mllm_dsa_prompt}
\end{figure*}

\section{Model Information}
\label{appendix:model_info}

For \textit{HSA} and \textit{DSA}, model generations used default temperature settings; for \textit{SI}, the temperature \textit{t} is zero when possible and near-zero (0.0001) if model configurations required $t$ to be a positive float. MiniCPM-V2.6, Qwen2-7B text-only base model, Qwen2-VL, Qwen2.5-VL, and Qwen2.5-Omni required near-zero temperatures. Information about standard-scale MLLMs is summarized in Table \ref{tab:model_info} to inform analysis of how training and architecture  influence social understanding.

\begin{table}[t]
\centering
\setlength{\tabcolsep}{1pt}
\renewcommand{\arraystretch}{1.1}
\small
\begin{tabular}{l|c|c|c}
\toprule
\rowcolor{header}
\multirow{2}{*}{\textbf{Model}} &
\multirow{2}{*}{\textbf{Video Modeling}} &
\multicolumn{2}{c}{\textbf{Vid. Data}} \\
\cmidrule(l){3-4}
\rowcolor{header}
\rule{0pt}{1em} & \textbf{Strategy} & \textbf{PT} & \textbf{FT} \\
\midrule
LongVA & UniRes - extended images & No & No \\
Oryx-1.5 & OryxVIT encoder + compression & No & Yes \\
LLaVA-NV & Image-as-frames & Yes & Yes \\
MiniCPM-V2.6 & Lightweight video encoder & No & No \\
Qwen2-VL & Image encoder & No & Yes \\
Qwen2.5-VL & Frame sequence + dynamic FPS & Yes & Yes \\
Qwen2.5-Omni & Video + audio encoder fusion & Yes & Yes \\
InternVL2 & Image encoder & No & Yes \\
InternVL3 & Video encoder w/ temporal fusion & No & Yes \\
\bottomrule
\end{tabular}
\caption{\textbf{Overview of Standard-Scale Video Language Models (7B–8B)}, comparing video-modeling strategies and the use of video data during pretraining (PT) and fine-tuning (FT). These features highlight architectural choices relevant to video understanding.}
\label{tab:model_info}
\end{table} 

\section{Subset Selection Random Sampling for Human Evaluation (N=50)}
\label{appendix:subset_selection}

To construct the 50-video subset used for human evaluation, we employ a stratified random sampling strategy designed to closely approximate the distribution of the full dataset across several observable axes: (1) \textit{SI difficulty}, where each video is assigned a difficulty level (very hard, hard, medium, easy) based on its average \textit{SI} accuracy across all models, (2) \textit{number of people per frame}, computed as the average number of visible individuals per frame and discretized into bins (1, 2, 3, 4+), (3) \textit{dialogue density}, measured as number of conversation turns per minute and grouped into 4 quartiles, (4) \textit{question category}, reflecting the question type based on the interrogative word used, i.e. \textit{who}, \textit{why}, \textit{what}, \textit{how}, and others. We preserve the per-model \textit{SI} accuracy distribution of the full dataset within the 50-video subset to avoid bias toward easier or harder samples.

As shown in Tables~\ref{tab:si_dist}--\ref{tab:model_dist}, the selected subset aligns with the full dataset across dimensions. Most distributional differences are within 1--2\%, ensuring that the subset remains representative of the overall data. This careful matching enables reliable and unbiased comparison of model performance across both \textit{HSA} and \textit{DSA} dimensions.
Table~\ref{tab:model_dist} shows that \texttt{Qwen2.5-Omni} is an outlier with a full-versus-subset \textit{SI} discrepancy, while the remaining evaluated models have closely matched full and subset \textit{SI} performance. We treat this result as a model-specific outlier. We inspected \texttt{Qwen2.5-Omni}'s open-ended \textit{HSA} and \textit{DSA} generations and observed cases of under-specified, repetitive, and malformed outputs. We retain these outputs, since generation reliability is part of model performance in open-ended social analysis. We discuss this information to contextualize the \texttt{Qwen2.5-Omni} results in \textsc{Social Caption}, not as a general claim beyond this domain.

\begin{table}[h]
\centering
\small
\setlength{\tabcolsep}{4pt}
\begin{tabular}{l|ccc}
\toprule
\rowcolor{header}
\textbf{SI Difficulty} & \textbf{Full} & \textbf{Subset} & \textbf{Abs. Diff} \\
\midrule
Very Hard & 0.255 & 0.240 & 0.015 \\
Hard      & 0.283 & 0.280 & 0.003 \\
Medium    & 0.214 & 0.220 & 0.006 \\
Easy      & 0.248 & 0.260 & 0.012 \\
\bottomrule
\end{tabular}
\caption{\textbf{SI difficulty distribution.} Proportion of videos at each difficulty level in the full validation set and the 50-video subset. The  absolute differences confirm that the subset closely mirrors the full distribution.}
\label{tab:si_dist}
\end{table}

\begin{table}[h]
\centering
\small
\setlength{\tabcolsep}{4pt}
\begin{tabular}{l|ccc}
\toprule
\rowcolor{header}
\textbf{People / Frame} & \textbf{Full} & \textbf{Subset} & \textbf{Abs. Diff} \\
\midrule
1   & 0.042 & 0.080 & 0.038 \\
2   & 0.503 & 0.480 & 0.023 \\
3   & 0.266 & 0.260 & 0.006 \\
4+  & 0.189 & 0.180 & 0.009 \\
\bottomrule
\end{tabular}
\caption{\textbf{Number of people per frame distribution.} Proportion of videos at each people-per-frame bin in the full dataset and the 50-video subset. The absolute differences confirm that the subset closely mirrors the full distribution.}
\label{tab:frame_dist}
\end{table}

\begin{table}[h]
\centering
\small
\setlength{\tabcolsep}{4pt}
\begin{tabular}{l|ccc}
\toprule
\rowcolor{header}
\textbf{Dialogue Density} & \textbf{Full} & \textbf{Subset} & \textbf{Abs. Diff} \\
\midrule
Low       & 0.255 & 0.240 & 0.015 \\
Medium    & 0.255 & 0.260 & 0.005 \\
High      & 0.248 & 0.260 & 0.012 \\
Very High & 0.241 & 0.240 & 0.001 \\
\bottomrule
\end{tabular}
\caption{\textbf{Dialogue density distribution.} Proportion of videos at each dialogue density level in the full dataset and the 50-video subset. The absolute differences confirm that the subset closely mirrors the full distribution.}
\label{tab:dialogue_dist}
\end{table}

\begin{table}[h]
\centering
\small
\setlength{\tabcolsep}{3pt}
\begin{tabular}{l|ccc}
\toprule
\rowcolor{header}
\textbf{Category} & \textbf{Full} & \textbf{Subset} & \textbf{Abs. Diff} \\
\midrule
Who      & 0.022 & 0.016 & 0.007 \\
Why      & 0.329 & 0.320  & 0.009 \\
Does     & 0.287 & 0.282 & 0.005 \\
Describe & 0.016 & 0.013 & 0.003 \\
How      & 0.066 & 0.060  & 0.006 \\
What     & 0.145 & 0.154 & 0.008 \\
Is       & 0.072 & 0.088 & 0.016 \\
Other    & 0.063 & 0.069 & 0.006 \\
\bottomrule
\end{tabular}
\caption{\textbf{Question category distribution.} Proportion of questions in each category across the full dataset and the 50-video subset. The  absolute differences confirm that the subset closely mirrors the full distribution.}
\label{tab:category_dist}
\end{table}

\begin{table}[h]
\centering
\small
\setlength{\tabcolsep}{3pt}
\begin{tabular}{l|ccc}
\toprule
\rowcolor{header}
\textbf{Model} & \textbf{Full (\%)} & \textbf{Subset (\%)} & \textbf{Abs. Diff (\%)} \\
\midrule
LLaVA-NV        & 68.94 & 68.99 & 0.05 \\
MiniCPM-V2.6      & 69.86 & 69.51 & 0.35 \\
Qwen2-VL     & 72.02 & 73.10 & 1.08 \\
Qwen2.5-VL   & 70.00 & 71.20 & 1.20 \\
Qwen2.5-Omni & 68.42 & 54.11 & 14.31 \\
InternVL2    & 67.77 & 68.04 & 0.27 \\
InternVL3    & 73.83 & 73.73 & 0.10 \\
GPT-4o       & 78.58 & 80.49 & 1.91 \\
Gemini-1.5-Pro   & 78.40 & 79.85 & 1.45 \\
Gemini-2.5-Pro   & 82.08 & 83.97 & 1.89 \\
\bottomrule
\end{tabular}
\caption{\textbf{Per-model \textit{SI} accuracy (\%) on the full dataset versus the representative 50-video subset.} For 9 of 10 models, absolute differences remain $\leq$1.91\%. The outlier \texttt{Qwen2.5-Omni} is discussed in Appendix \ref{appendix:subset_selection}.}
\label{tab:model_dist}
\end{table}

\section{Power Analysis and Sample Size} 
\label{appendix:sample-comparison}

For human evaluation of \textit{HSA} and \textit{DSA}, 50 videos were randomly sampled to ensure representation of diverse video contexts. Each MLLM processed these 50 videos and generated \textit{HSA} and \textit{DSA} descriptions; two human annotators rated these descriptions using a 6-criterion Likert rubric on a scale of 1-5. 
Human evaluation required approximately 100 hours of annotator effort, ensuring high-quality and reliable judgments. In total, the study includes 12,000 annotations (6,000 per \textit{HSA} and \textit{DSA} dimension). A paired t-test power analysis indicates that with $n=50$ videos we achieve 80\% power to detect effects of  $d_z \ge 0.40$. The 50-video subset closely mirrors the full validation set in model \textit{SI} performance and was jointly matched across SI difficulty, social complexity, dialogue density, and question-category distributions, supporting the treatment of the sampled videos as a representative subset (see Appendix \ref{appendix:subset_selection} for details).

\section{Inter-Annotator Agreement Analysis}
\label{appendix:IAA}
To assess annotator agreement, we computed three complementary metrics: the binary F1 score, the binary agreement score, and the average absolute difference between Likert-scale ratings. To obtain binary labels, each 5-point Likert score was mapped to a binary value: scores of 3 and above were assigned as 1 (high quality), and scores below 3 as 0 (low quality). The first two metrics capture the proportion of matching judgments based on binary labels. The binary agreement score measures the percentage of instances where both annotators assigned the same binary label, providing an indication of categorical agreement. The binary F1 score balances precision and recall to account for cases where one annotator may systematically rate higher or lower than the other, offering a more nuanced view of alignment on positive (high-quality) judgments. The third metric, the average absolute difference, measures the mean difference between annotators’ original 5-point Likert scores, reflecting how close ratings are on a continuous scale; smaller values of average absolute difference indicate higher consistency and stronger  agreement.

Compared to measures such as Cohen’s Kappa or Pearson’s $r$, the metrics that we selected are intended to more reliably assess agreement within our annotation setup. Videos were evaluated by distinct pairs of annotators on Prolific. 

\begin{table*}[t]
  \centering
  \setlength{\tabcolsep}{5pt}
  \renewcommand{\arraystretch}{1.2}
  \small
  \begin{tabular}{lccc|ccc}
    \toprule
    \rowcolor{header}
    \multirow{2}{*}{\textbf{Comparison}} & 
    \multicolumn{3}{c|}{\textbf{Holistic Social Analysis}} & 
    \multicolumn{3}{c}{\textbf{Directed Social Analysis}} \\
    \cmidrule(lr){2-4} \cmidrule(lr){5-7}
    \rowcolor{header}
    & \makecell{\textbf{Binary F1} \\ \textbf{Score (\%)}} 
    & \makecell{\textbf{Binary Agreement} \\ \textbf{Score (\%)}} 
    & \makecell{\textbf{Avg. Abs.} \\ \textbf{Difference}} 
    & \makecell{\textbf{Binary F1} \\ \textbf{Score (\%)}} 
    & \makecell{\textbf{Binary Agreement} \\ \textbf{Score (\%)}} 
    & \makecell{\textbf{Avg. Abs.} \\ \textbf{Difference}}\\
    \midrule
    H-H            & 86.60 & 62.00 & \colorbox{lowlight}{1.07}\textcolor{low}{$\downarrow$} & 84.87 & \colorbox{lowlight}{63.20}\textcolor{low}{$\downarrow$} & \colorbox{lowlight}{1.10}\textcolor{low}{$\downarrow$} \\
    H-G   & \colorbox{lowlight}{86.47}\textcolor{low}{$\downarrow$} & \colorbox{highlight}{\textbf{63.47}}\textcolor{high}{$\uparrow$} & 1.05 & \colorbox{lowlight}{84.50}\textcolor{low}{$\downarrow$} & \colorbox{highlight}{\textbf{69.83}}\textcolor{high}{$\uparrow$} & 1.04 \\
    H-I8B      & 91.44 & \colorbox{lowlight}{58.47}\textcolor{low}{$\downarrow$} & 1.06 & 89.50 & 65.70 & 0.90 \\
    H-I78B     & \textbf{\colorbox{highlight}{91.69}}\textcolor{high}{$\uparrow$} & 60.70 & \textbf{\colorbox{highlight}{0.93}}\textcolor{high}{$\uparrow$} & \textbf{\colorbox{highlight}{90.26}}\textcolor{high}{$\uparrow$} & 68.10 & \textbf{\colorbox{highlight}{0.84}}\textcolor{high}{$\uparrow$} \\
    \bottomrule
  \end{tabular}
  \caption{\textbf{Combined Inter-Annotator Agreement for \textit{HSA} and \textit{DSA} Dimensions.} H-H: Human--Human, H-G: Human--Gemini-2.5-Pro, H-I8B: Human--InternVL3-8B, H-I78B: Human--InternVL3-78B. Binary F1 Score (\%) quantifies overlap between annotations, Binary Agreement Score (\%) captures categorical alignment, and Average Absolute Difference (Avg. Abs.) measures rating deviation. \textcolor{high}{$\uparrow$} Highest and \textcolor{low}{$\downarrow$} Lowest performance per metric.}
  \label{tab:combined_agreement}
\end{table*}

Table~\ref{tab:combined_agreement} presents the inter-annotator agreement for both \textit{HSA} and \textit{DSA} dimensions. Table \ref{tab:binary_agreement} and Table ~\ref{tab:combined_binary_diff} present a fine-grained analysis of inter-annotator and model–human agreement across the six evaluation criteria for \textit{HSA} and \textit{DSA}. 

\begin{table*}[t]
\centering
\footnotesize
\renewcommand{\arraystretch}{0.75}
\setlength{\tabcolsep}{2pt}
\begin{tabular}{p{2.4cm}|cccc|cccc}
\toprule
\rowcolor{header}\multirow{2}{*}{\textbf{Sub-Dimensions}} & 
    \multicolumn{4}{c|}{\cellcolor{header}\textbf{Binary Agreement Score (\%)}} &
    \multicolumn{4}{c}{\cellcolor{header}\textbf{Average Absolute Difference}} \\
\cmidrule(lr){2-5} \cmidrule(l){6-9}
\rowcolor{header} 
 & \textbf{H-H} & \textbf{H-G} & \textbf{H-I8B} & \textbf{H-I78B} 
 & \textbf{H-H} & \textbf{H-G} & \textbf{H-I8B} & \textbf{H-I78B} \\
\midrule
\multicolumn{9}{l}{\textbf{HSA}} \\
Scene Desc.        & \textbf{\colorbox{highlight}{64.4}}\textcolor{high}{$\uparrow$} & 63.4 & \colorbox{lowlight}{61.0}\textcolor{low}{$\downarrow$} & 63.4 & \colorbox{lowlight}{0.99}\textcolor{low}{$\downarrow$} & 0.92 & 0.98 & \textbf{\colorbox{highlight}{0.82}}\textcolor{high}{$\uparrow$} \\
Individuals        & 60.4 & \textbf{\colorbox{highlight}{60.6}}\textcolor{high}{$\uparrow$} & \colorbox{lowlight}{52.8}\textcolor{low}{$\downarrow$} & 56.2 & 1.17 & 1.11 & \colorbox{lowlight}{1.18}\textcolor{low}{$\downarrow$} & \textbf{\colorbox{highlight}{0.98}}\textcolor{high}{$\uparrow$} \\
Topic \& Context      & 60.2 & \textbf{\colorbox{highlight}{64.2}}\textcolor{high}{$\uparrow$} & \colorbox{lowlight}{56.2}\textcolor{low}{$\downarrow$} & 58.6 & \colorbox{lowlight}{1.12}\textcolor{low}{$\downarrow$} & 1.08 & 1.07 & \textbf{\colorbox{highlight}{0.97}}\textcolor{high}{$\uparrow$} \\
Socio-Emotional    & \colorbox{lowlight}{63.6}\textcolor{low}{$\downarrow$} & 66.4 & 66.0 & \textbf{\colorbox{highlight}{67.6}}\textcolor{high}{$\uparrow$} & \colorbox{lowlight}{1.09}\textcolor{low}{$\downarrow$} & 1.06 & 1.01 & \textbf{\colorbox{highlight}{0.87}}\textcolor{high}{$\uparrow$} \\
Answer Det.        & \textbf{\colorbox{highlight}{62.0}}\textcolor{high}{$\uparrow$} & \textbf{\colorbox{highlight}{62.0}}\textcolor{high}{$\uparrow$} & \colorbox{lowlight}{55.6}\textcolor{low}{$\downarrow$} & 56.8 & 1.03 & 1.05 & \colorbox{lowlight}{1.11}\textcolor{low}{$\downarrow$} & \textbf{\colorbox{highlight}{0.94}}\textcolor{high}{$\uparrow$} \\
Prompt Adher.      & 61.4 & \textbf{\colorbox{highlight}{64.2}}\textcolor{high}{$\uparrow$} & \colorbox{lowlight}{59.2}\textcolor{low}{$\downarrow$} & 61.6 & 1.05 & \colorbox{lowlight}{1.06}\textcolor{low}{$\downarrow$} & \textbf{\colorbox{highlight}{1.03}}\textcolor{high}{$\uparrow$} & 1.04 \\
\midrule
\multicolumn{9}{l}{\textbf{DSA}} \\
Rel. Scene Det.  & \colorbox{lowlight}{68.8}\textcolor{low}{$\downarrow$} & 70.2 & \colorbox{lowlight}{68.8}\textcolor{low}{$\downarrow$} & \textbf{\colorbox{highlight}{70.8}}\textcolor{high}{$\uparrow$} & \colorbox{lowlight}{1.01}\textcolor{low}{$\downarrow$} & 0.88 & 0.86 & \textbf{\colorbox{highlight}{0.79}}\textcolor{high}{$\uparrow$} \\
Key Individuals  & \colorbox{lowlight}{59.8}\textcolor{low}{$\downarrow$} & \textbf{\colorbox{highlight}{70.2}}\textcolor{high}{$\uparrow$} & 63.0 & 65.8 & \colorbox{lowlight}{1.16}\textcolor{low}{$\downarrow$} & 1.01 & 0.92 & \textbf{\colorbox{highlight}{0.84}}\textcolor{high}{$\uparrow$} \\
Rel. Interact.   & \colorbox{lowlight}{62.0}\textcolor{low}{$\downarrow$} & \textbf{\colorbox{highlight}{71.0}}\textcolor{high}{$\uparrow$} & 67.0 & 69.8 & \colorbox{lowlight}{1.11}\textcolor{low}{$\downarrow$} & 1.08 & 0.87 & \textbf{\colorbox{highlight}{0.83}}\textcolor{high}{$\uparrow$} \\
Rel. Context     & 65.8 & \textbf{\colorbox{highlight}{70.2}}\textcolor{high}{$\uparrow$} & \colorbox{lowlight}{65.6}\textcolor{low}{$\downarrow$} & 68.4 & \colorbox{lowlight}{1.06}\textcolor{low}{$\downarrow$} & 1.05 & 0.91 & \textbf{\colorbox{highlight}{0.85}}\textcolor{high}{$\uparrow$} \\
Answer Det.      & \colorbox{lowlight}{61.0}\textcolor{low}{$\downarrow$} & \textbf{\colorbox{highlight}{68.6}}\textcolor{high}{$\uparrow$} & 66.6 & 66.8 & 1.14 & \colorbox{lowlight}{1.21}\textcolor{low}{$\downarrow$} & 0.90 & \textbf{\colorbox{highlight}{0.87}}\textcolor{high}{$\uparrow$} \\
Prompt Adher.    & \colorbox{lowlight}{61.8}\textcolor{low}{$\downarrow$} & \textbf{\colorbox{highlight}{68.8}}\textcolor{high}{$\uparrow$} & 63.2 & 67.0 & \colorbox{lowlight}{1.10}\textcolor{low}{$\downarrow$} & 1.02 & 0.93 & \textbf{\colorbox{highlight}{0.84}}\textcolor{high}{$\uparrow$} \\
\bottomrule
\end{tabular}
\caption{\textbf{Binary Agreement Score and Average Absolute Difference for \textit{HSA} and \textit{DSA} Sub-Dimensions.} H-H: Human--Human, H-G: Human--Gemini-2.5-Pro, H-I8B: Human--InternVL3-8B, H-I78B: Human--InternVL3-78B. \textcolor{high}{$\uparrow$} Highest and \textcolor{low}{$\downarrow$} Lowest performance per metric.}
\label{tab:combined_binary_diff}
\end{table*}

\section{Model Size Variants and Social Inference Performance}
\label{appendix:model_size_variants}

This section presents the \textit{SI} performance of different size variants of InternVL3 and Qwen2-VL models, highlighting the impact of model scale and the inclusion of transcriptions. Table~\ref{tab:internvl3_qwen2_size_variant} reports the \textit{SI} accuracy for each model variant both with and without transcription input. These results are also visualized in Figure~\ref{fig:size_variants} in the main paper.

\begin{table}[ht]
\centering
\small
\renewcommand{\arraystretch}{1.1}
\setlength{\tabcolsep}{3pt}
\begin{tabular}{ccc|cc}
\toprule
{\cellcolor{header}\textbf{Size}} &
\cellcolor{header}\textbf{CL} &
{\cellcolor{header}\textbf{Backbone}} &
\multicolumn{2}{c}{\cellcolor{header}\textbf{Social Inference (\%)}} \\
\cmidrule(lr){4-5}
\rowcolor{header}
& & &
\textbf{w/o trans.} & \textbf{w/ trans.} \\
\midrule
\multicolumn{5}{l}{\textbf{InternVL3 Size Variants}} \\
\midrule
1B  & 32K & Qwen2.5-0.5B  & \colorbox{lowlight}{50.85}\textcolor{low}{$\downarrow$} & \colorbox{lowlight}{53.72}\textcolor{low}{$\downarrow$} \\
2B  & 32K & Qwen2.5-1.5B  & 63.30 & 65.75 \\
8B  & 32K & Qwen2.5-7B  & 67.23 & 73.83 \\
9B  & 32K & InternLM3-8B  & 66.60 & 71.49 \\
14B & 32K & Qwen2.5-14B & 70.96 & 75.53 \\
38B & 32K & Qwen2.5-32B & \textbf{\colorbox{highlight}{73.62}}\textcolor{high}{$\uparrow$} & 78.19 \\
78B & 32K & Qwen2.5-72B & 72.77 & \textbf{\colorbox{highlight}{78.72}}\textcolor{high}{$\uparrow$} \\
\midrule
\multicolumn{5}{l}{\textbf{Qwen2-VL Size Variants}} \\
\midrule
2B  & 32K & Qwen2-2B  & \colorbox{lowlight}{58.09}\textcolor{low}{$\downarrow$} & \colorbox{lowlight}{65.21}\textcolor{low}{$\downarrow$} \\
7B  & 32K & Qwen2-7B  & 65.32 & 72.02 \\
72B & 32K & Qwen2-72B & \textbf{\colorbox{highlight}{73.30}}\textcolor{high}{$\uparrow$} & \textbf{\colorbox{highlight}{79.04}}\textcolor{high}{$\uparrow$} \\
\bottomrule
\end{tabular}
\caption{\textbf{InternVL3 and Qwen2-VL Size Variants: \textit{SI} Performance.}
Scores are percentages for evaluation with and without transcription.
\textcolor{high}{$\uparrow$} and \textcolor{low}{$\downarrow$} mark the highest and lowest values within each family.}
\label{tab:internvl3_qwen2_size_variant}
\end{table}

\section{Evaluation Results for Human and MLLM Judges}
\label{appendix:eval}

We performed both human and automated evaluations using Gemini-2.5-Pro, InternVL3-8B, and InternVL3-78B as MLLM judges on a subset of 50 videos. Each social understanding sub-dimension was independently evaluated by two human annotators, and the reported human evaluation scores represent the mean across annotators, averaged over all videos. The same set of sub-dimensions was also evaluated by the MLLM judges to generate automated assessment scores for comparison. Table~\ref{tab:numbers_hsa} presents the human and MLLM-based evaluation results for the \textit{HSA} dimension, and Table~\ref{tab:numbers_dsa} shows the corresponding results for the \textit{DSA} dimension.

\begin{table*}[t]
  \centering
  \setlength{\tabcolsep}{2.5pt}
  \renewcommand{\arraystretch}{1.2}
  \scriptsize
  \begin{tabular}{@{}l | c|c|c|c|c|c|c@{}}
    \toprule
    \rowcolor{header}
    \textbf{Model} & 
    \makecell{\textbf{Scene}\\\textbf{Desc. (5)}} & 
    \makecell{\textbf{Individuals}\\\textbf{(5)}} & 
    \makecell{\textbf{Topic \&}\\\textbf{Context (5)}} & 
    \makecell{\textbf{Socio-}\\\textbf{Emotional (5)}} & 
    \makecell{\textbf{Answer}\\\textbf{Detail (5)}} & 
    \makecell{\textbf{Prompt Adherence}\\\textbf{(5)}} & 
    \makecell{\textbf{Total}\\\textbf{(30)}} \\
    \specialrule{1.2pt}{2pt}{2pt}
    \multicolumn{8}{l}{\textbf{Human Evaluation}} \\
    \specialrule{1.2pt}{2pt}{2pt}
    LLaVA-NV            & 3.90 & 3.67 & 3.49 & 3.59 & 3.44 & 3.63 & 21.72 \\
    MiniCPM-V2.6         & 3.81 & 3.46 & 3.77 & \textbf{\colorbox{highlight}{3.93}}\textcolor{high}{$\uparrow$} & 3.64 & 3.74 & 22.35 \\
    Qwen2-VL              & 3.89 & 3.49 & 3.42 & 3.50 & 3.54 & 3.79 & 21.63 \\
    Qwen2.5-VL           & \textbf{\colorbox{highlight}{4.09}}\textcolor{high}{$\uparrow$} & 3.67 & \textbf{\colorbox{highlight}{3.85}}\textcolor{high}{$\uparrow$} & 3.87 & \textbf{\colorbox{highlight}{3.83}}\textcolor{high}{$\uparrow$} & 3.93 & \textbf{\colorbox{highlight}{23.24}}\textcolor{high}{$\uparrow$} \\
    Qwen2.5-Omni         & \colorbox{lowlight}{2.73}\textcolor{low}{$\downarrow$} & \colorbox{lowlight}{2.57}\textcolor{low}{$\downarrow$} & \colorbox{lowlight}{2.54}\textcolor{low}{$\downarrow$} & \colorbox{lowlight}{2.37}\textcolor{low}{$\downarrow$} & \colorbox{lowlight}{2.36}\textcolor{low}{$\downarrow$} & \colorbox{lowlight}{2.27}\textcolor{low}{$\downarrow$} & \colorbox{lowlight}{14.84}\textcolor{low}{$\downarrow$} \\
    InternVL2             & 4.07 & \textbf{\colorbox{highlight}{3.69}}\textcolor{high}{$\uparrow$} & \textbf{\colorbox{highlight}{3.85}}\textcolor{high}{$\uparrow$} & 3.72 & 3.79 & \textbf{\colorbox{highlight}{4.01}}\textcolor{high}{$\uparrow$} & 23.13 \\
    InternVL3             & 3.74 & 3.49 & 3.53 & 3.70 & 3.52 & 3.80 & 21.78 \\
    \midrule
    GPT-4o                & \colorbox{lowlight}{3.73}\textcolor{low}{$\downarrow$} & \colorbox{lowlight}{3.72}\textcolor{low}{$\downarrow$} & \colorbox{lowlight}{3.79}\textcolor{low}{$\downarrow$} & \colorbox{lowlight}{3.99}\textcolor{low}{$\downarrow$} & \colorbox{lowlight}{3.64}\textcolor{low}{$\downarrow$} & \colorbox{lowlight}{3.98}\textcolor{low}{$\downarrow$} & \colorbox{lowlight}{22.85}\textcolor{low}{$\downarrow$} \\
    Gemini-1.5-Pro        & 4.09 & 3.96 & 4.07 & 4.14 & 4.06 & 4.14 & 24.46 \\
    Gemini-2.5-Pro        & \textbf{\colorbox{highlight}{4.33}}\textcolor{high}{$\uparrow$} & \textbf{\colorbox{highlight}{4.37}}\textcolor{high}{$\uparrow$} & \textbf{\colorbox{highlight}{4.35}}\textcolor{high}{$\uparrow$} & \textbf{\colorbox{highlight}{4.38}}\textcolor{high}{$\uparrow$} & \textbf{\colorbox{highlight}{4.47}}\textcolor{high}{$\uparrow$} & \textbf{\colorbox{highlight}{4.24}}\textcolor{high}{$\uparrow$} & \textbf{\colorbox{highlight}{26.14}}\textcolor{high}{$\uparrow$} \\
    \bottomrule
    \specialrule{1.2pt}{2pt}{2pt}
    \multicolumn{8}{l}{\textbf{MLLM-as-Judge Evaluation - Gemini-2.5-Pro}} \\
    \specialrule{1.2pt}{2pt}{2pt}
    LLaVA-NV & 4.30 & 3.12 & 3.16 & 2.82 & 2.48 & 4.74 & 20.62 \\
    MiniCPM-V2.6    & 4.44 & 3.16 & \textbf{\colorbox{highlight}{3.82}}\textcolor{high}{$\uparrow$} & 3.54 & 3.04 & 4.94 & 22.94 \\
    Qwen2-VL        & 4.02 & 2.74 & 3.04 & 2.88 & 2.54 & 4.94 & 20.16 \\
    Qwen2.5-VL      & \textbf{\colorbox{highlight}{4.62}}\textcolor{high}{$\uparrow$} & \textbf{\colorbox{highlight}{3.40}}\textcolor{high}{$\uparrow$} & 3.66 & \textbf{\colorbox{highlight}{3.58}}\textcolor{high}{$\uparrow$} & \textbf{\colorbox{highlight}{3.40}}\textcolor{high}{$\uparrow$} & \textbf{\colorbox{highlight}{4.96}}\textcolor{high}{$\uparrow$} & \textbf{\colorbox{highlight}{23.62}}\textcolor{high}{$\uparrow$} \\
    Qwen2.5-Omni    & \colorbox{lowlight}{3.06}\textcolor{low}{$\downarrow$} & \colorbox{lowlight}{2.30}\textcolor{low}{$\downarrow$} & \colorbox{lowlight}{2.14}\textcolor{low}{$\downarrow$} & \colorbox{lowlight}{1.80}\textcolor{low}{$\downarrow$} & \colorbox{lowlight}{1.82}\textcolor{low}{$\downarrow$} & \colorbox{lowlight}{1.94}\textcolor{low}{$\downarrow$} & \colorbox{lowlight}{13.06}\textcolor{low}{$\downarrow$} \\
    InternVL2       & 4.12 & 2.90 & 3.56 & 3.40 & 2.88 & 4.92 & 21.78 \\
    InternVL3       & 4.58 & 3.38 & 3.74 & \textbf{\colorbox{highlight}{3.58}}\textcolor{high}{$\uparrow$} & 3.16 & 4.94 & 23.38 \\
\midrule
GPT-4o          & \colorbox{lowlight}{4.82}\textcolor{low}{$\downarrow$} & 4.30 & 4.68 & \colorbox{lowlight}{4.46}\textcolor{low}{$\downarrow$} & \colorbox{lowlight}{4.28}\textcolor{low}{$\downarrow$} & \textbf{\colorbox{highlight}{5.00}}\textcolor{high}{$\uparrow$} & 27.54 \\
Gemini-1.5-Pro  & 4.88 & \colorbox{lowlight}{3.98}\textcolor{low}{$\downarrow$} & \colorbox{lowlight}{4.34}\textcolor{low}{$\downarrow$} & 4.48 & 4.32 & \colorbox{lowlight}{4.98}\textcolor{low}{$\downarrow$} & \colorbox{lowlight}{26.98}\textcolor{low}{$\downarrow$} \\
Gemini-2.5-Pro  & \textbf{\colorbox{highlight}{5.00}}\textcolor{high}{$\uparrow$} & \textbf{\colorbox{highlight}{4.96}}\textcolor{high}{$\uparrow$} & \textbf{\colorbox{highlight}{5.00}}\textcolor{high}{$\uparrow$} & \textbf{\colorbox{highlight}{5.00}}\textcolor{high}{$\uparrow$} & \textbf{\colorbox{highlight}{5.00}}\textcolor{high}{$\uparrow$} & \textbf{\colorbox{highlight}{5.00}}\textcolor{high}{$\uparrow$} & \textbf{\colorbox{highlight}{29.96}}\textcolor{high}{$\uparrow$} \\
\bottomrule
\specialrule{1.2pt}{2pt}{2pt}
        \rowcolor{header}
        \textbf{Model} & 
        \makecell{\textbf{Scene}\\\textbf{Desc. (5)}} & 
        \makecell{\textbf{Individuals}\\\textbf{(5)}} & 
        \makecell{\textbf{Topic \&}\\\textbf{Context (5)}} & 
        \makecell{\textbf{Socio-}\\\textbf{Emotional (5)}} & 
        \makecell{\textbf{Answer}\\\textbf{Detail (5)}} & 
        \makecell{\textbf{Adherence}\\\textbf{(5)}} & 
        \makecell{\textbf{Total}\\\textbf{(30)}} \\
        \specialrule{1.2pt}{2pt}{2pt}
        \multicolumn{8}{l}{\textbf{MLLM-as-Judge Evaluation - InternVL3-8B}} \\
        \specialrule{1.2pt}{2pt}{2pt}
        LLaVA-NV & 4.68 & 4.66 & 4.66 & 4.64 & 4.40 & 4.68 & 27.72 \\
        MiniCPM-V2.6    & 4.88 & 4.88 & 4.88 & 4.88 & 4.78 & 4.88 & 29.18 \\
        Qwen2-VL         & 4.44 & 4.42 & 4.40 & 4.38 & 4.02 & 4.46 & 26.12 \\
        Qwen2.5-VL      & 4.88 & 4.88 & 4.88 & 4.88 & 4.84 & 4.90 & 29.26 \\
        Qwen2.5-Omni    & \colorbox{lowlight}{3.96}\textcolor{low}{$\downarrow$} & \colorbox{lowlight}{3.86}\textcolor{low}{$\downarrow$} & \colorbox{lowlight}{3.58}\textcolor{low}{$\downarrow$} & \colorbox{lowlight}{3.36}\textcolor{low}{$\downarrow$} & \colorbox{lowlight}{3.26}\textcolor{low}{$\downarrow$} & \colorbox{lowlight}{3.72}\textcolor{low}{$\downarrow$} & \colorbox{lowlight}{21.74}\textcolor{low}{$\downarrow$} \\
        InternVL2        & 4.70 & 4.68 & 4.64 & 4.64 & 4.44 & 4.66 & 27.76 \\
        InternVL3        & \textbf{\colorbox{highlight}{4.94}}\textcolor{high}{$\uparrow$} & \textbf{\colorbox{highlight}{4.92}}\textcolor{high}{$\uparrow$} & \textbf{\colorbox{highlight}{4.92}}\textcolor{high}{$\uparrow$} & \textbf{\colorbox{highlight}{4.92}}\textcolor{high}{$\uparrow$} & \textbf{\colorbox{highlight}{4.92}}\textcolor{high}{$\uparrow$} & \textbf{\colorbox{highlight}{4.94}}\textcolor{high}{$\uparrow$} & \textbf{\colorbox{highlight}{29.56}}\textcolor{high}{$\uparrow$} \\
        \midrule
        GPT-4o           & \colorbox{lowlight}{4.86}\textcolor{low}{$\downarrow$} & \colorbox{lowlight}{4.86}\textcolor{low}{$\downarrow$} & \colorbox{lowlight}{4.86}\textcolor{low}{$\downarrow$} & \colorbox{lowlight}{4.86}\textcolor{low}{$\downarrow$} & \colorbox{lowlight}{4.80}\textcolor{low}{$\downarrow$} & \colorbox{lowlight}{4.92}\textcolor{low}{$\downarrow$} & \colorbox{lowlight}{29.16}\textcolor{low}{$\downarrow$} \\
        Gemini-1.5-Pro   & 4.96 & 4.96 & 4.96 & 4.96 & 4.96 & 4.98 & 29.78 \\
        Gemini-2.5-Pro   & \textbf{\colorbox{highlight}{5.00}}\textcolor{high}{$\uparrow$} & \textbf{\colorbox{highlight}{5.00}}\textcolor{high}{$\uparrow$} & \textbf{\colorbox{highlight}{5.00}}\textcolor{high}{$\uparrow$} & \textbf{\colorbox{highlight}{5.00}}\textcolor{high}{$\uparrow$} & \textbf{\colorbox{highlight}{5.00}}\textcolor{high}{$\uparrow$} & \textbf{\colorbox{highlight}{5.00}}\textcolor{high}{$\uparrow$} & \textbf{\colorbox{highlight}{30.00}}\textcolor{high}{$\uparrow$} \\
        \bottomrule
        \specialrule{1.2pt}{2pt}{2pt}
        \multicolumn{8}{l}{\textbf{MLLM-as-Judge Evaluation - InternVL3-78B}} \\
        \specialrule{1.2pt}{2pt}{2pt}
LLaVA-NV & 4.44 & 4.30 & 4.36 & 4.26 & 4.14 & 4.72 & 26.22 \\
MiniCPM-V2.6     & 4.54 & 4.50 & 4.66 & 4.48 & 4.48 & 4.88 & 27.54 \\
Qwen2-VL         & 4.16 & 3.88 & 4.28 & 4.04 & 3.86 & 4.62 & 24.84 \\
Qwen2.5-VL      & 4.54 & 4.50 & 4.54 & 4.50 & 4.50 & 4.90 & 27.48 \\
Qwen2.5-Omni    & \colorbox{lowlight}{3.56}\textcolor{low}{$\downarrow$} & 
                   \colorbox{lowlight}{3.28}\textcolor{low}{$\downarrow$} & 
                   \colorbox{lowlight}{3.16}\textcolor{low}{$\downarrow$} & 
                   \colorbox{lowlight}{2.88}\textcolor{low}{$\downarrow$} & 
                   \colorbox{lowlight}{3.02}\textcolor{low}{$\downarrow$} & 
                   \colorbox{lowlight}{3.42}\textcolor{low}{$\downarrow$} & 
                   \colorbox{lowlight}{19.32}\textcolor{low}{$\downarrow$} \\
InternVL2        & 4.26 & 4.08 & 4.32 & 4.24 & 4.12 & 4.70 & 25.72 \\
InternVL3        & \textbf{\colorbox{highlight}{4.76}}\textcolor{high}{$\uparrow$} & \textbf{\colorbox{highlight}{4.70}}\textcolor{high}{$\uparrow$} & \textbf{\colorbox{highlight}{4.82}}\textcolor{high}{$\uparrow$} & \textbf{\colorbox{highlight}{4.76}}\textcolor{high}{$\uparrow$} & \textbf{\colorbox{highlight}{4.70}}\textcolor{high}{$\uparrow$} & \textbf{\colorbox{highlight}{4.94}}\textcolor{high}{$\uparrow$} & \textbf{\colorbox{highlight}{28.68}}\textcolor{high}{$\uparrow$} \\
\midrule
GPT-4o           & \colorbox{lowlight}{4.60}\textcolor{low}{$\downarrow$} & \colorbox{lowlight}{4.60}\textcolor{low}{$\downarrow$} & \colorbox{lowlight}{4.78}\textcolor{low}{$\downarrow$} & \colorbox{lowlight}{4.70}\textcolor{low}{$\downarrow$} & \colorbox{lowlight}{4.66}\textcolor{low}{$\downarrow$} & \textbf{\colorbox{highlight}{5.00}}\textcolor{high}{$\uparrow$} & \colorbox{lowlight}{28.34}\textcolor{low}{$\downarrow$} \\
Gemini-1.5-Pro   & 4.64 & \colorbox{lowlight}{4.60}\textcolor{low}{$\downarrow$} & 4.82 & 4.72 & \colorbox{lowlight}{4.66}\textcolor{low}{$\downarrow$} & \colorbox{lowlight}{4.94}\textcolor{low}{$\downarrow$} & 28.38 \\
Gemini-2.5-Pro   & \textbf{\colorbox{highlight}{4.98}}\textcolor{high}{$\uparrow$} & 
                   \textbf{\colorbox{highlight}{4.98}}\textcolor{high}{$\uparrow$} & 
                   \textbf{\colorbox{highlight}{5.00}}\textcolor{high}{$\uparrow$} & 
                   \textbf{\colorbox{highlight}{5.00}}\textcolor{high}{$\uparrow$} & 
                   \textbf{\colorbox{highlight}{5.00}}\textcolor{high}{$\uparrow$} & 
                   \textbf{\colorbox{highlight}{5.00}}\textcolor{high}{$\uparrow$} & 
                   \textbf{\colorbox{highlight}{29.96}}\textcolor{high}{$\uparrow$} \\
        \bottomrule
  \end{tabular}
  \caption{\textbf{HSA Model Performance.}
Scores reported across 6 sub-dimensions (max score of 5 each, max total score of 30) under Human Evaluation and three MLLM-as-Judge settings (Gemini-2.5-Pro, InternVL3-8B, InternVL3-78B). \textcolor{high}{$\uparrow$} Highest and \textcolor{low}{$\downarrow$} Lowest scores per column are highlighted for open-source and closed-source models.}
  \label{tab:numbers_hsa}
\end{table*}

\begin{table*}[ht]
    \centering
  \setlength{\tabcolsep}{2.5pt}
  \renewcommand{\arraystretch}{1.15}
  \scriptsize
  \begin{tabular}{@{}l | c|c|c|c|c|c|c@{}}
    \toprule
    \rowcolor{header}
    \textbf{Model} & 
    \makecell{\textbf{Relevant}\\\textbf{Scene Details (5)}} & 
    \makecell{\textbf{Key}\\\textbf{Individuals (5)}} & 
    \makecell{\textbf{Relevant}\\\textbf{Interactions (5)}} & 
    \makecell{\textbf{Relevant}\\\textbf{Context (5)}} & 
    \makecell{\textbf{Answer}\\\textbf{Detail (5)}} & 
    \makecell{\textbf{Prompt}\\\textbf{Adher. (5)}} & 
    \makecell{\textbf{Total}\\\textbf{(30)}} \\
    \specialrule{1.2pt}{2pt}{2pt}
    \multicolumn{8}{l}{\textbf{Human Evaluation}} \\
    \specialrule{1.2pt}{2pt}{2pt}
LLaVA-NV       & 3.46 & 3.49 & 3.29 & 3.31 & 3.22 & 3.37 & 20.14 \\
MiniCPM-V2.6     & 4.02 & \textbf{\colorbox{highlight}{3.90}}\textcolor{high}{$\uparrow$} & \textbf{\colorbox{highlight}{3.87}}\textcolor{high}{$\uparrow$} & \textbf{\colorbox{highlight}{3.93}}\textcolor{high}{$\uparrow$} & 3.80 & 3.99 & \textbf{\colorbox{highlight}{23.51}}\textcolor{high}{$\uparrow$} \\
Qwen2-VL         & 3.30 & 3.11 & 2.97 & 3.15 & 2.97 & 3.40 & 18.90 \\
Qwen2.5-VL      & \textbf{\colorbox{highlight}{4.23}}\textcolor{high}{$\uparrow$} & 3.78 & 3.69 & 3.83 & \textbf{\colorbox{highlight}{3.93}}\textcolor{high}{$\uparrow$} & \textbf{\colorbox{highlight}{4.03}}\textcolor{high}{$\uparrow$} & 23.49 \\
Qwen2.5-Omni    & \colorbox{lowlight}{1.39}\textcolor{low}{$\downarrow$} & \colorbox{lowlight}{1.42}\textcolor{low}{$\downarrow$} & \colorbox{lowlight}{1.45}\textcolor{low}{$\downarrow$} & \colorbox{lowlight}{1.49}\textcolor{low}{$\downarrow$} & \colorbox{lowlight}{1.52}\textcolor{low}{$\downarrow$} & \colorbox{lowlight}{1.64}\textcolor{low}{$\downarrow$} & \colorbox{lowlight}{8.91}\textcolor{low}{$\downarrow$} \\
InternVL2        & 3.53 & 3.30 & 3.29 & 3.34 & 3.12 & 3.30 & 19.88 \\
InternVL3        & 4.05 & 3.85 & 3.69 & 3.79 & 3.72 & 3.91 & 23.01 \\
    \midrule
    GPT-4o           & 4.12 & 4.05 & 4.12 & 4.04 & 3.92 & 4.17 & 24.42 \\
Gemini-1.5-Pro   & \colorbox{lowlight}{3.85}\textcolor{low}{$\downarrow$} & \colorbox{lowlight}{3.86}\textcolor{low}{$\downarrow$} & \colorbox{lowlight}{3.87}\textcolor{low}{$\downarrow$} & \colorbox{lowlight}{3.95}\textcolor{low}{$\downarrow$} & \colorbox{lowlight}{3.91}\textcolor{low}{$\downarrow$} & \colorbox{lowlight}{3.98}\textcolor{low}{$\downarrow$} & \colorbox{lowlight}{23.42}\textcolor{low}{$\downarrow$} \\
Gemini-2.5-Pro   & \textbf{\colorbox{highlight}{4.24}}\textcolor{high}{$\uparrow$} & \textbf{\colorbox{highlight}{4.27}}\textcolor{high}{$\uparrow$} & \textbf{\colorbox{highlight}{4.38}}\textcolor{high}{$\uparrow$} & \textbf{\colorbox{highlight}{4.38}}\textcolor{high}{$\uparrow$} & \textbf{\colorbox{highlight}{4.37}}\textcolor{high}{$\uparrow$} & \textbf{\colorbox{highlight}{4.30}}\textcolor{high}{$\uparrow$} & \textbf{\colorbox{highlight}{25.94}}\textcolor{high}{$\uparrow$} \\

    \bottomrule
    \specialrule{1.2pt}{2pt}{2pt}
    \multicolumn{8}{l}{\textbf{MLLM-as-Judge Evaluation - Gemini-2.5-Pro}} \\
        \specialrule{1.2pt}{2pt}{2pt}
            LLaVA-NV & 3.72 & 2.92 & 2.28 & 2.46 & 1.98 & 3.86 & 17.22 \\
MiniCPM-V2.6    & 4.56 & 3.48 & 3.10 & \textbf{\colorbox{highlight}{3.94}}\textcolor{high}{$\uparrow$} & 3.02 & 4.88 & 22.98 \\
Qwen2-VL        & 3.34 & 2.56 & 2.26 & 2.36 & 2.06 & 3.74 & 16.32 \\
Qwen2.5-VL     & \textbf{\colorbox{highlight}{4.68}}\textcolor{high}{$\uparrow$} & 3.54 & \textbf{\colorbox{highlight}{3.30}}\textcolor{high}{$\uparrow$} & 3.72 & 2.96 & \textbf{\colorbox{highlight}{4.96}}\textcolor{high}{$\uparrow$} & \textbf{\colorbox{highlight}{23.16}}\textcolor{high}{$\uparrow$} \\
Qwen2.5-Omni   & \colorbox{lowlight}{1.18}\textcolor{low}{$\downarrow$} & \colorbox{lowlight}{1.18}\textcolor{low}{$\downarrow$} & \colorbox{lowlight}{1.06}\textcolor{low}{$\downarrow$} & \colorbox{lowlight}{1.12}\textcolor{low}{$\downarrow$} & \colorbox{lowlight}{1.20}\textcolor{low}{$\downarrow$} & \colorbox{lowlight}{1.14}\textcolor{low}{$\downarrow$} & \colorbox{lowlight}{6.88}\textcolor{low}{$\downarrow$} \\
InternVL2       & 3.30 & 2.22 & 1.98 & 2.50 & 1.94 & 4.28 & 16.22 \\
InternVL3       & 4.60 & \textbf{\colorbox{highlight}{3.82}}\textcolor{high}{$\uparrow$} & 3.22 & 3.38 & \textbf{\colorbox{highlight}{3.06}}\textcolor{high}{$\uparrow$} & 4.90 & 22.98 \\

        \midrule
        GPT-4o           & 4.94 & 4.38 & 4.18 & \colorbox{lowlight}{4.24}\textcolor{low}{$\downarrow$} & \colorbox{lowlight}{4.00}\textcolor{low}{$\downarrow$} & \textbf{\colorbox{highlight}{4.96}}\textcolor{high}{$\uparrow$} & 26.70 \\
        Gemini-1.5-Pro   & \colorbox{lowlight}{4.74}\textcolor{low}{$\downarrow$} & \colorbox{lowlight}{4.10}\textcolor{low}{$\downarrow$} & \colorbox{lowlight}{4.02}\textcolor{low}{$\downarrow$} & 4.48 & 4.18 & \colorbox{lowlight}{4.90}\textcolor{low}{$\downarrow$} & \colorbox{lowlight}{26.42}\textcolor{low}{$\downarrow$} \\
        Gemini-2.5-Pro   & \textbf{\colorbox{highlight}{4.96}}\textcolor{high}{$\uparrow$} & \textbf{\colorbox{highlight}{4.88}}\textcolor{high}{$\uparrow$} & \textbf{\colorbox{highlight}{4.82}}\textcolor{high}{$\uparrow$} & \textbf{\colorbox{highlight}{4.88}}\textcolor{high}{$\uparrow$} & \textbf{\colorbox{highlight}{4.88}}\textcolor{high}{$\uparrow$} & \textbf{\colorbox{highlight}{4.96}}\textcolor{high}{$\uparrow$} & \textbf{\colorbox{highlight}{29.38}}\textcolor{high}{$\uparrow$} \\
        \bottomrule
        \specialrule{1.2pt}{2pt}{2pt}
        \rowcolor{header}
    \textbf{Model} & 
    \makecell{\textbf{Rel.}\\\textbf{Scene Det. (5)}} & 
    \makecell{\textbf{Key}\\\textbf{Individuals (5)}} & 
    \makecell{\textbf{Rel.}\\\textbf{Interactions (5)}} & 
    \makecell{\textbf{Rel.}\\\textbf{Context (5)}} & 
    \makecell{\textbf{Answer}\\\textbf{Det. (5)}} & 
    \makecell{\textbf{Prompt}\\\textbf{Adher. (5)}} & 
    \makecell{\textbf{Total}\\\textbf{(30)}} \\
        \specialrule{1.2pt}{2pt}{2pt}
    \multicolumn{8}{l}{\textbf{MLLM-as-Judge Evaluation - InternVL3-8B}} \\
        \specialrule{1.2pt}{2pt}{2pt}
        LLaVA-NV       & 4.32 & 4.30 & 4.14 & 4.18 & 4.00 & 4.48 & 25.42 \\
        MiniCPM-V2.6    & 4.66 & 4.66 & 4.54 & \textbf{\colorbox{highlight}{4.64}}\textcolor{high}{$\uparrow$} & 4.56 & 4.80 & 27.86 \\
        Qwen2-VL         & 3.94 & 3.98 & 3.60 & 3.74 & 3.60 & 4.28 & 23.14 \\
        Qwen2.5-VL      & 4.62 & 4.62 & 4.54 & 4.56 & 4.56 & 4.78 & 27.68 \\
        Qwen2.5-Omni    & \colorbox{lowlight}{2.76}\textcolor{low}{$\downarrow$} & 
                           \colorbox{lowlight}{2.66}\textcolor{low}{$\downarrow$} & 
                           \colorbox{lowlight}{1.98}\textcolor{low}{$\downarrow$} & 
                           \colorbox{lowlight}{2.14}\textcolor{low}{$\downarrow$} & 
                           \colorbox{lowlight}{1.96}\textcolor{low}{$\downarrow$} & 
                           \colorbox{lowlight}{2.36}\textcolor{low}{$\downarrow$} & 
                           \colorbox{lowlight}{13.86}\textcolor{low}{$\downarrow$} \\
        InternVL2        & 3.90 & 3.82 & 3.24 & 3.58 & 3.36 & 3.94 & 21.84 \\
        InternVL3        & \textbf{\colorbox{highlight}{4.68}}\textcolor{high}{$\uparrow$} & \textbf{\colorbox{highlight}{4.68}}\textcolor{high}{$\uparrow$} & \textbf{\colorbox{highlight}{4.58}}\textcolor{high}{$\uparrow$} & \textbf{\colorbox{highlight}{4.64}}\textcolor{high}{$\uparrow$} & \textbf{\colorbox{highlight}{4.62}}\textcolor{high}{$\uparrow$} & \textbf{\colorbox{highlight}{4.88}}\textcolor{high}{$\uparrow$} & \textbf{\colorbox{highlight}{28.08}}\textcolor{high}{$\uparrow$} \\
        \midrule
        GPT-4o           & 4.70 & 4.70 & 4.62 & 4.68 & 4.68 & 4.90 & 28.28 \\
        Gemini-1.5-Pro   & \colorbox{lowlight}{4.36}\textcolor{low}{$\downarrow$} & \colorbox{lowlight}{4.42}\textcolor{low}{$\downarrow$} & \colorbox{lowlight}{4.24}\textcolor{low}{$\downarrow$} & \colorbox{lowlight}{4.26}\textcolor{low}{$\downarrow$} & \colorbox{lowlight}{4.36}\textcolor{low}{$\downarrow$} & \colorbox{lowlight}{4.72}\textcolor{low}{$\downarrow$} & \colorbox{lowlight}{26.36}\textcolor{low}{$\downarrow$} \\
        Gemini-2.5-Pro   & \textbf{\colorbox{highlight}{4.92}}\textcolor{high}{$\uparrow$} & 
                           \textbf{\colorbox{highlight}{4.94}}\textcolor{high}{$\uparrow$} & 
                           \textbf{\colorbox{highlight}{4.86}}\textcolor{high}{$\uparrow$} & 
                           \textbf{\colorbox{highlight}{4.92}}\textcolor{high}{$\uparrow$} & 
                           \textbf{\colorbox{highlight}{4.92}}\textcolor{high}{$\uparrow$} & 
                           \textbf{\colorbox{highlight}{4.94}}\textcolor{high}{$\uparrow$} & 
                           \textbf{\colorbox{highlight}{29.50}}\textcolor{high}{$\uparrow$} \\
        \bottomrule
        \specialrule{1.2pt}{2pt}{2pt}
    \multicolumn{8}{l}{\textbf{MLLM-as-Judge Evaluation - InternVL3-78B}} \\
        \specialrule{1.2pt}{2pt}{2pt}
      LLaVA-NV       & 4.22 & 4.16 & 3.86 & 4.10 & 3.94 & 4.32 & 24.60 \\
        MiniCPM-V2.6    & 4.46 & 4.40 & 4.34 & 4.44 & 4.40 & 4.66 & 26.70 \\
        Qwen2-VL         & 3.78 & 3.66 & 3.46 & 3.62 & 3.52 & 4.02 & 22.06 \\
        Qwen2.5-VL      & 4.54 & 4.54 & 4.40 & 4.54 & 4.44 & 4.70 & 27.16 \\
        Qwen2.5-Omni    & \colorbox{lowlight}{2.02}\textcolor{low}{$\downarrow$} & \colorbox{lowlight}{1.96}\textcolor{low}{$\downarrow$} & \colorbox{lowlight}{1.88}\textcolor{low}{$\downarrow$} & \colorbox{lowlight}{1.90}\textcolor{low}{$\downarrow$} & \colorbox{lowlight}{1.90}\textcolor{low}{$\downarrow$} & \colorbox{lowlight}{2.00}\textcolor{low}{$\downarrow$} & \colorbox{lowlight}{11.66}\textcolor{low}{$\downarrow$} \\
        InternVL2        & 3.82 & 3.60 & 3.24 & 3.58 & 3.28 & 3.98 & 21.50 \\
        InternVL3        & \textbf{\colorbox{highlight}{4.68}}\textcolor{high}{$\uparrow$} & \textbf{\colorbox{highlight}{4.64}}\textcolor{high}{$\uparrow$} & \textbf{\colorbox{highlight}{4.56}}\textcolor{high}{$\uparrow$} & \textbf{\colorbox{highlight}{4.62}}\textcolor{high}{$\uparrow$} & \textbf{\colorbox{highlight}{4.54}}\textcolor{high}{$\uparrow$} & \textbf{\colorbox{highlight}{4.80}}\textcolor{high}{$\uparrow$} & \textbf{\colorbox{highlight}{27.84}}\textcolor{high}{$\uparrow$} \\
        \midrule
        GPT-4o           & 4.54 & 4.54 & 4.50 & 4.54 & 4.52 & \textbf{\colorbox{highlight}{4.86}}\textcolor{high}{$\uparrow$} & 27.50 \\
        Gemini-1.5-Pro   & \colorbox{lowlight}{4.16}\textcolor{low}{$\downarrow$} & \colorbox{lowlight}{4.16}\textcolor{low}{$\downarrow$} & \colorbox{lowlight}{3.98}\textcolor{low}{$\downarrow$} & \colorbox{lowlight}{4.10}\textcolor{low}{$\downarrow$} & \colorbox{lowlight}{4.12}\textcolor{low}{$\downarrow$} & \colorbox{lowlight}{4.38}\textcolor{low}{$\downarrow$} & \colorbox{lowlight}{24.90}\textcolor{low}{$\downarrow$} \\
        Gemini-2.5-Pro   & \textbf{\colorbox{highlight}{4.60}}\textcolor{high}{$\uparrow$} & \textbf{\colorbox{highlight}{4.68}}\textcolor{high}{$\uparrow$} & \textbf{\colorbox{highlight}{4.64}}\textcolor{high}{$\uparrow$} & \textbf{\colorbox{highlight}{4.68}}\textcolor{high}{$\uparrow$} & \textbf{\colorbox{highlight}{4.70}}\textcolor{high}{$\uparrow$} & 4.70 & \textbf{\colorbox{highlight}{28.00}}\textcolor{high}{$\uparrow$} \\
        \bottomrule
    \end{tabular}
    \caption{\textbf{DSA Model Performance.}
Scores reported across 6 sub-dimensions (max score of 5 each, max total score of 30) under Human Evaluation and three MLLM-as-Judge settings (Gemini-2.5-Pro, InternVL3-8B, InternVL3-78B). \textcolor{high}{$\uparrow$} Highest and \textcolor{low}{$\downarrow$} Lowest scores per column are highlighted for open-source and closed-source models.}
    \label{tab:numbers_dsa}
\end{table*}

\section{Mismatch Test}
\label{sec:baseline_scores}

We conducted a mismatch test to assess whether MLLM judges are assigning  evaluation scores indiscriminately to model-generated \textit{HSA} and \textit{DSA}  responses. For each model, we paired its \textit{HSA} and \textit{DSA} response with three unrelated videos. Each MLLM judge evaluated a pair of unrelated video and model response. Low mismatch scores indicate that a judge penalizes model generations that do not semantically align with social information in videos, rather than assigning high scores based upon surface-level aspects of generation. We averaged the scores across the three mismatched pairings to obtain a mismatch baseline for each judge and model pair. Tables~\ref{tab:mllm_hsa_baseline_combined} and~\ref{tab:mllm_dsa_combined} show the results of this test for our three MLLM judges evaluating 10 models across both \textit{HSA} and \textit{DSA} sub-dimensions.

\begin{table*}[t]
  \centering
  \setlength{\tabcolsep}{5pt} 
  \renewcommand{\arraystretch}{1.2} 
  \small
  \begin{tabular}{@{}l | c|c|c|c|c|c|c@{}}
    \toprule
    \rowcolor{header}
        \textbf{Model} & 
        \makecell{\textbf{Scene}\\\textbf{Desc. (5)}} & 
        \makecell{\textbf{Individuals}\\\textbf{(5)}} & 
        \makecell{\textbf{Topic \&}\\\textbf{Context (5)}} & 
        \makecell{\textbf{Socio-}\\\textbf{Emotional (5)}} & 
        \makecell{\textbf{Answer}\\\textbf{Detail (5)}} & 
        \makecell{\textbf{Prompt Adherence}\\\textbf{(5)}} & 
        \makecell{\textbf{Total}\\\textbf{(30)}} \\
    \specialrule{1.2pt}{2pt}{2pt}
    \multicolumn{8}{l}{\textbf{Baseline Scores – MLLM-as-Judge (Gemini-2.5-Pro)}} \\
    \specialrule{1.2pt}{2pt}{2pt}
    LLaVA-NV   & 1.05 & 1.01 & 1.00 & 1.03 & 1.00 & 4.57 & 9.66 \\
    MiniCPM-V2.6       & 1.01 & 1.00 & 1.00 & 1.03 & 1.00 & 4.74 & 9.78 \\
    Qwen2-VL           & 1.01 & 1.00 & 1.00 & 1.05 & 1.00 & 4.78 & 9.84 \\
    Qwen2.5-VL         & 1.01 & 1.01 & 1.00 & 1.02 & 1.00 & 4.57 & 9.61 \\
    Qwen2.5-Omni       & 1.00 & 1.00 & 1.00 & 1.01 & 1.00 & 1.82 & 6.83 \\
    InternVL2          & 1.01 & 1.00 & 1.01 & 1.07 & 1.00 & 4.75 & 9.84 \\
    InternVL3          & 1.10 & 1.01 & 1.00 & 1.05 & 1.00 & 4.66 & 9.82 \\
    GPT-4o             & 1.08 & 1.02 & 1.00 & 1.03 & 1.00 & 4.80 & 9.93 \\
    Gemini-1.5-Pro     & 1.00 & 1.00 & 1.00 & 1.00 & 1.00 & 4.31 & 9.31 \\
    Gemini-2.5-Pro     & 1.00 & 1.00 & 1.00 & 1.00 & 1.00 & 4.13 & 9.13 \\
    \bottomrule
    \specialrule{1.2pt}{2pt}{2pt}
    \multicolumn{8}{l}{\textbf{Baseline Scores – MLLM-as-Judge (InternVL3-8B)}} \\
    \specialrule{1.2pt}{2pt}{2pt}
    LLaVA-NV  & 2.16 & 1.85 & 1.79 & 1.77 & 1.73 & 1.63 & 10.93 \\
    MiniCPM-V2.6      & 2.46 & 2.08 & 2.05 & 1.97 & 1.94 & 1.89 & 12.39 \\
    Qwen2-VL          & 2.38 & 2.06 & 1.93 & 1.89 & 1.85 & 1.81 & 11.92 \\
    Qwen2.5-VL        & 2.86 & 2.45 & 2.39 & 2.29 & 2.13 & 2.22 & 14.34 \\
    Qwen2.5-Omni      & 2.17 & 1.82 & 1.68 & 1.65 & 1.59 & 1.54 & 10.45 \\
    InternVL2         & 2.21 & 1.95 & 1.77 & 1.75 & 1.73 & 1.73 & 11.14 \\
    InternVL3         & 2.36 & 2.01 & 1.93 & 1.87 & 1.80 & 1.82 & 11.79 \\
    GPT-4o            & 2.28 & 2.03 & 1.91 & 1.85 & 1.81 & 1.73 & 11.61 \\
    Gemini-1.5-Pro    & 2.29 & 1.99 & 1.90 & 1.83 & 1.76 & 1.60 & 11.37 \\
    Gemini-2.5-Pro    & 2.57 & 2.13 & 2.05 & 1.96 & 1.83 & 1.71 & 12.25 \\
    \bottomrule
    \specialrule{1.2pt}{2pt}{2pt}
    \multicolumn{8}{l}{\textbf{Baseline Scores – MLLM-as-Judge (InternVL3-78B)}} \\
    \specialrule{1.2pt}{2pt}{2pt}
    LLaVA-NV  & 1.26 & 1.23 & 1.27 & 1.25 & 1.25 & 1.27 & 7.53 \\
    MiniCPM-V2.6      & 1.47 & 1.41 & 1.43 & 1.40 & 1.43 & 1.46 & 8.60 \\
    Qwen2-VL          & 1.38 & 1.33 & 1.37 & 1.35 & 1.35 & 1.38 & 8.16 \\
    Qwen2.5-VL        & 1.38 & 1.35 & 1.37 & 1.33 & 1.35 & 1.39 & 8.17 \\
    Qwen2.5-Omni      & 1.14 & 1.12 & 1.11 & 1.11 & 1.11 & 1.13 & 6.72 \\
    InternVL2         & 1.40 & 1.35 & 1.40 & 1.39 & 1.39 & 1.40 & 8.33 \\
    InternVL3         & 1.39 & 1.35 & 1.29 & 1.33 & 1.37 & 1.39 & 8.12 \\
    GPT-4o            & 1.35 & 1.31 & 1.33 & 1.31 & 1.32 & 1.35 & 7.97 \\
    Gemini-1.5-Pro    & 1.30 & 1.29 & 1.29 & 1.28 & 1.29 & 1.30 & 7.75 \\
    Gemini-2.5-Pro    & 1.23 & 1.21 & 1.24 & 1.21 & 1.21 & 1.23 & 7.33 \\
    \bottomrule
  \end{tabular}
  \caption{\textbf{Baseline Scores Across Models for HSA with MLLM-as-Judge}\\
      Baseline scores from MLLM-as-Judge evaluations with Gemini-2.5-Pro, InternVL3-8B, and InternVL3-78B.}
  \label{tab:mllm_hsa_baseline_combined}
\end{table*}

\begin{table*}[t]
  \centering
  \setlength{\tabcolsep}{4pt} 
  \renewcommand{\arraystretch}{1.2} 
  \small
  \begin{tabular}{@{}l | c|c|c|c|c|c|c@{}}
    \toprule
    \rowcolor{header}
    \textbf{Model} & \makecell{\textbf{Relevant}\\\textbf{Scene Det. (5)}} & 
    \makecell{\textbf{Key}\\\textbf{Individuals (5)}} & 
    \makecell{\textbf{Relevant}\\\textbf{Interactions (5)}} & 
    \makecell{\textbf{Relevant}\\\textbf{Context (5)}} & 
    \makecell{\textbf{Answer}\\\textbf{Det. (5)}} & 
    \makecell{\textbf{Prompt}\\\textbf{Adherence (5)}} & 
    \makecell{\textbf{Total}\\\textbf{(30)}} \\
    \midrule
    
    \multicolumn{8}{l}{\textbf{Baseline Scores – MLLM-as-Judge (Gemini-2.5-Pro)}} \\
    \midrule
    LLaVA-NV & 1.05 & 1.01 & 1.00 & 1.00 & 1.01 & 2.05 & 7.12 \\
    MiniCPM-V2.6    & 1.01 & 1.00 & 1.01 & 1.00 & 1.00 & 1.97 & 6.99 \\
    Qwen2-VL         & 1.00 & 1.00 & 1.01 & 1.01 & 1.01 & 2.09 & 7.12 \\
    Qwen2.5-VL       & 1.02 & 1.00 & 1.01 & 1.01 & 1.00 & 2.04 & 7.08 \\
    Qwen2.5-Omni     & 1.00 & 1.00 & 1.00 & 1.00 & 1.01 & 1.05 & 6.06 \\
    InternVL2        & 1.01 & 1.00 & 1.01 & 1.00 & 1.00 & 2.38 & 7.40 \\
    InternVL3        & 1.00 & 1.00 & 1.03 & 1.00 & 1.00 & 2.40 & 7.43 \\
    GPT-4o           & 1.05 & 1.00 & 1.00 & 1.00 & 1.00 & 2.56 & 7.61 \\
    Gemini-1.5-Pro   & 1.01 & 1.00 & 1.00 & 1.00 & 1.00 & 1.27 & 6.28 \\
    Gemini-2.5-Pro   & 1.00 & 1.00 & 1.00 & 1.00 & 1.00 & 1.23 & 6.23 \\
    \midrule
    \multicolumn{8}{l}{\textbf{Baseline Scores – MLLM-as-Judge (InternVL3-8B)}} \\
    \midrule
    LLaVA-NV  & 1.97 & 1.79 & 1.47 & 1.41 & 1.25 & 1.39 & 9.28 \\
    MiniCPM-V2.6     & 1.87 & 1.76 & 1.39 & 1.40 & 1.25 & 1.24 & 8.91 \\
    Qwen2-VL          & 1.97 & 1.82 & 1.43 & 1.42 & 1.18 & 1.24 & 9.06 \\
    Qwen2.5-VL        & 2.01 & 1.93 & 1.52 & 1.48 & 1.35 & 1.40 & 9.69 \\
    Qwen2.5-Omni      & 2.55 & 2.03 & 1.68 & 1.87 & 1.38 & 1.71 & 11.22 \\
    InternVL2         & 1.95 & 1.88 & 1.33 & 1.35 & 1.15 & 1.24 & 8.90 \\
    InternVL3         & 1.93 & 1.87 & 1.41 & 1.43 & 1.26 & 1.30 & 9.20 \\
    GPT-4o            & 1.92 & 1.81 & 1.32 & 1.28 & 1.14 & 1.13 & 8.60 \\
    Gemini-1.5-Pro    & 1.98 & 1.72 & 1.43 & 1.38 & 1.18 & 1.20 & 8.89 \\
    Gemini-2.5-Pro    & 1.90 & 1.59 & 1.31 & 1.29 & 1.20 & 1.21 & 8.50 \\
    \midrule
    \multicolumn{8}{l}{\textbf{Baseline Scores – MLLM-as-Judge (InternVL3-78B)}} \\
    \midrule
    LLaVA-NV  & 1.23 & 1.23 & 1.21 & 1.21 & 1.22 & 1.23 & 7.33 \\
    MiniCPM-V2.6     & 1.17 & 1.17 & 1.16 & 1.16 & 1.16 & 1.18 & 7.00 \\
    Qwen2-VL          & 1.15 & 1.15 & 1.13 & 1.13 & 1.14 & 1.14 & 6.84 \\
    Qwen2.5-VL        & 1.18 & 1.18 & 1.18 & 1.17 & 1.17 & 1.18 & 7.06 \\
    Qwen2.5-Omni      & 1.07 & 1.06 & 1.05 & 1.06 & 1.04 & 1.05 & 6.33 \\
    InternVL2         & 1.25 & 1.23 & 1.21 & 1.21 & 1.22 & 1.23 & 7.35 \\
    InternVL3         & 1.20 & 1.20 & 1.19 & 1.19 & 1.19 & 1.20 & 7.17 \\
    GPT-4o            & 1.21 & 1.19 & 1.19 & 1.19 & 1.19 & 1.21 & 7.18 \\
    Gemini-1.5-Pro    & 1.10 & 1.10 & 1.09 & 1.09 & 1.09 & 1.09 & 6.56 \\
    Gemini-2.5-Pro    & 1.15 & 1.14 & 1.14 & 1.14 & 1.14 & 1.14 & 6.85 \\
    \bottomrule
  \end{tabular}
  \caption{\textbf{Baseline Scores Across Models for DSA with MLLM-as-Judge.} Baseline scores from MLLM-as-Judge evaluations with Gemini, InternVL3-8B, and InternVL3-78B.}
  \label{tab:mllm_dsa_combined}
\end{table*}

\end{document}